\documentclass{article}

    \PassOptionsToPackage{numbers, compress}{natbib}
\usepackage[preprint]{neurips_2023}

\usepackage[utf8]{inputenc} % allow utf-8 input
\usepackage[T1]{fontenc}    % use 8-bit T1 fonts
\usepackage{hyperref}       % hyperlinks
\usepackage{url}            % simple URL typesetting
\usepackage{booktabs}       % professional-quality tables
\usepackage{amsfonts}       % blackboard math symbols
\usepackage{nicefrac}       % compact symbols for 1/2, etc.
\usepackage{microtype}      % microtypography
\usepackage{xcolor}         % colors
\usepackage{xspace}
\usepackage{graphicx}
\usepackage{amsmath}
\usepackage{multirow}
\usepackage{tabularx, makecell}

\usepackage{listings} % for code block
\usepackage{tikz}
\usepackage{enumitem}
\usepackage{colortbl}
\usepackage{tcolorbox}

\definecolor{Gray}{gray}{0.5}
\definecolor{LGray}{gray}{0.9}
\definecolor{darkblue}{RGB}{94,110,186}
\definecolor{darkGreen}{RGB}{92,148,110}
\definecolor{myblue}{RGB}{14, 121, 178}

\newcommand{\violet}[1]{\textcolor{violet}{#1}}

\newcommand{\dataname}{InternVid}
\newcommand{\modelname}{ViCLIP}

\title{\dataname: A Large-scale Video-Text Dataset for Multimodal Understanding and Generation}
\author{%
  \hspace{-0.25cm}\textbf{Yi Wang$^{*1}$, Yinan He$^{*1}$, Yizhuo Li$^{*4,1}$, Kunchang Li$^{6,1}$, Jiashuo Yu$^{1}$, Xin Ma$^{3,1}$, Xinhao Li$^{2,1}$}\\
  \hspace{-0.25cm}\textbf{Guo Chen$^{3,1}$, Xinyuan Chen$^{1}$, Yaohui Wang$^{1}$, Conghui He$^{1}$, Ping Luo$^{4,1}$, Ziwei Liu$^{5,1}$}\\
  \hspace{-0.25cm}\textbf{Yali Wang$^{\dagger 6,1}$, Limin Wang$^{\dagger 2,1}$, Yu Qiao$^{\dagger 1}$}\\
 \hspace{-0.25cm}$^1$OpenGVLab, Shanghai AI Laboratory \quad $^2$Nanjing University \quad $^3$Monash University\\  \hspace{-0.25cm} $^4$The University of Hong Kong \quad $^5$Nanyang Technological University\\ 
 \hspace{-0.25cm}$^6$Shenzhen Institutes of Advanced Technology, Chinese Academy of Sciences \\
 \\
 {\url{https://github.com/OpenGVLab/InternVideo/tree/main/Data/InternVid}} \\
 \hspace{-0.25cm}
}

\begin{document}

\maketitle

\newcommand\blfootnote[1]{%
  \begingroup
  \renewcommand\thefootnote{}\footnote{#1}%
  \addtocounter{footnote}{-1}%
  \endgroup
}

\begin{abstract}
 This paper introduces {\bf {\dataname}}, a large-scale video-centric multimodal dataset that enables learning powerful and transferable video-text representations for multimodal understanding and generation.
 The {\bf {\dataname}} dataset contains over 7 million videos lasting nearly 760K hours, yielding 234M video clips accompanied by detailed descriptions of total 4.1B words. Our core contribution is to develop a scalable approach to autonomously build a high-quality video-text dataset with large language models (LLM), thereby showcasing its efficacy in learning video-language representation at scale.
 Specifically, we utilize a multi-scale approach to generate video-related descriptions. Furthermore, we introduce {\violet{\modelname}}, a video-text representation learning model based on ViT-L. Learned on {\dataname} via contrastive learning, this model demonstrates leading zero-shot action recognition and competitive video retrieval performance. Beyond basic video understanding tasks like recognition and retrieval, our dataset and model have broad applications. They are particularly beneficial for generating interleaved video-text data for learning a video-centric dialogue system, advancing video-to-text and text-to-video generation research. These proposed resources provide a tool for researchers and practitioners interested in multimodal video understanding and generation. %The data, code, and models are publicly available. 
 \blfootnote{* Equal contribution.\ \ \ \ $\dagger$  Corresponding authors.}
\end{abstract}

\begin{figure*}[ht]
    \centering
    \includegraphics[width=1\textwidth]{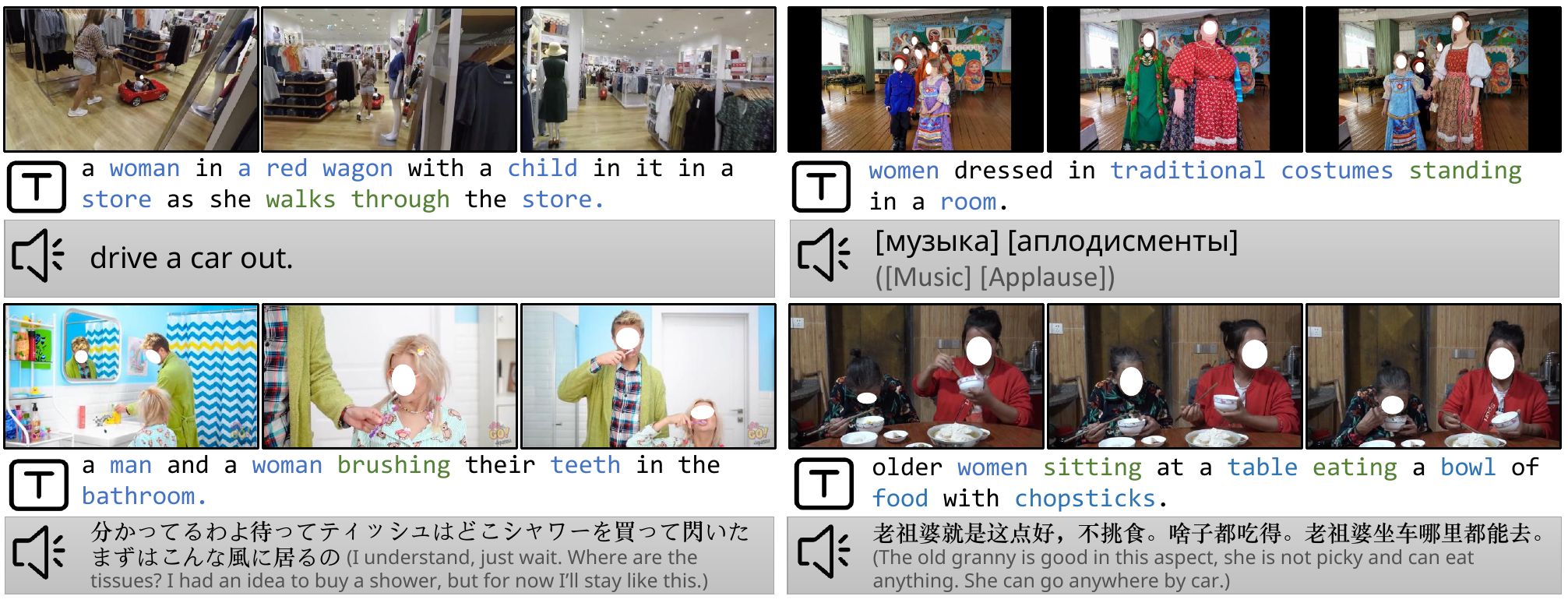}
    % \vspace{-0.3cm}
    \caption{Examples (we give three frames of each video clip), the corresponding generated captions, and ASR transcripts in {\dataname}. In the captions, we highlight nouns in \textcolor{myblue}{blue} and verbs in \textcolor{darkGreen}{green}. Non-English transcripts are translated to English using LLM \cite{brown2020gpt3}.
    }
    \label{fig:teaser}
    \vspace{-0.3cm}
\end{figure*}

\section{Introduction}
Learning transferable video-text representations is both challenging and essential for video understanding in various real-world applications such as autonomous driving, intelligent surveillance, human-computer interaction, and visual searching. While multimodal contrastive learning using web-scale data has been successful in image-text representation, it remains underexplored in the video-language domain.

A key reason for this limited exploration is \textit{the lack of a high quality video-language dataset for pretraining at scale}. Current research relies on datasets like HowTo100M \cite{miech2019howto100m}, HD-VILA \cite{xue2022advancing}, and YT-Temporal \cite{zellers2021merlot,zellers2022merlot}, whose texts are generated using automatic speech recognition (ASR). Despite their large scale, these datasets often have low semantic correlations between the videos and corresponding textual descriptions \cite{miech2019howto100m,xue2022advancing,zellers2021merlot,zellers2022merlot}. Empirical studies demonstrate that improving this correlation (e.g. aligning videos with subtitles to improve their matching) significantly benefits downstream tasks such as video retrieval and video question answering \cite{bain2021frozen}. Recent works have utilized WebVid10M \cite{bain2021frozen}, a dataset with higher-quality alt-texts, to address the low video-text correlation issue. However, its limited scale and dynamics hinder its use in current data and model scaling studies. Specifically, only 10M video-text pairs are provided, and the depicted scenes contain relatively few actions or activities. 

We propose a large-scale video-centric dataset {\dataname} to address the challenge of \textit{scaling up video-language modeling while maintaining high video-text correspondence}. Visual examples are given in Figure \ref{fig:teaser}. Note the ASR transcripts barely depict visual elements in videos while the generated captions do. The dataset contains highly-correlated video-text pairs and includes over 7 million videos, totaling 760,000 hours and resulting in 234 million video clips, with various subsets for different needs. These videos cover 16 scenarios and around 6,000 motion descriptions. To improve video-text matching, we generate captions using a multiscale approach. In the coarse scale, we caption the middle frame of each video and use the description as the video caption. In the fine scale, we produce frame-by-frame captions and summarize them with a language model. 
% Empirical evidence of the effectiveness of {\dataname} is presented in Section \ref{sec:ablation}.

Leveraging {\dataname}, we scale a video-language transformer (ViT-L) in contrastive learning from a data perspective, and its experiments prove {\dataname} enables learning scalable video-text models. We introduce video masking to the model to accelerate the whole learning without compromising its effectiveness. The video and text encoders are initialized from the CLIP pretrained model with the same scale. With {\dataname}, we learn a video-text model for several epochs, achieving impressive zero-shot performance. Compared with previous Video CLIP variants, our proposed {\modelname} shows notable performance improvement, especially in zero-shot settings.

In addition to large-scale video-language contrastive pretraining, we discover its effectiveness in producing interleaved video-text data for learning a video-centric dialogue system like Flamingo \cite{flamingo,openflamingo}, and advancing video generation. Since the text-annotated clips are extracted from videos, we naturally collect clips and their corresponding text based on the sampling locations. This results in approximately 7 million interleaved data pieces, suitable for instruction tuning as multi-turn video-centric dialogue. For video generation, we filter the core set and obtain 18 million video clips. Alongside WebVid-10M, {\dataname} can significantly improve a stable-diffusion based video generation model to new heights.

In summary, our contributions are threefold. 
\begin{itemize}[leftmargin=*]
    \item[$\bullet$] We introduce a new web-scale video-language dataset {\dataname}. This dataset, aimed at advancing video-related multimodal understanding and generation at scale, is created using a multi-scale video captioning approach powered by LLM, ensuring high-quality video-text data with minimal human intervention. {\dataname} has 7 million videos, corresponding to 234 million clips each with the generated captions. Spanning 16 scenes and about 6 thousand actions, the dataset includes computational features (video-text correlation and visual aesthetics) across the entirely of the dataset and gives way to diverse subsets to cater to varying training needs.
    \item[$\bullet$] We learn a new video-language model, {\modelname}, which is trained on {\dataname} using ViT-L. It incorporates both constrastive learning and mask modeling techniques, allowing for efficient learning of transferrable video-language representation. This model achieves state-of-the-art zero-shot action recognition in Kinetics, scoring 75.7, 73.5, and 66.4 on K400, K600, and K700 with the average top1 and top5 accuracies, respectively. It also gets competitive performance on video retrieval, setting a new baseline for video-text understanding.
    \item[$\bullet$] {\dataname} fosters the development of multimodal dialogue systems and text-to-video generation.
    The proposed {\modelname} learned on {\dataname} could serve as a vision backbone of video-centric dialogue systems\cite{zhu2023minigpt,li2023videochat,liu2023internchat}, conducting tasks as action recognition, temporal understanding, reasoning, and creativity within an open-ended environment. Furthermore, we provide a subset, {\dataname}-Aesthetics, created using specific video-text relation and visual aesthetic filtering. This subset aids in generating high-resolution watermark-free videos. Utilizing {\dataname}-Aesthetics, both visual and quantitative outcomes of a simple text-to-video baseline can be noticeably enhanced (FVD: 705.3 -> 616.5).
    % \item[$\bullet$] We conducted an empirical validation in {\dataname} and found that captioning models generate more effective video descriptions than alt-text for video-text understanding. Our analysis of existing training methods suggests that false negatives are amplified in the context of video-language learning. Notably, a diversified subset of {\dataname} (10M) outperforms the nearly full dataset (200M) significantly in zero-shot action recognition and video retrieval experiments.
    %\item[$\bullet$] We learn a new video-language model, {\modelname}, using ViT-L based on {\dataname}. This model achieves state-of-the-art zero-shot action recognition in Kinetics (75.7, 73.5, and 66.4 on K400, K600, and K700 with the average top1 and top5 accuracies, respectively). It also gets competitive performance on video retrieval, setting a new baseline for video-text understanding. With its strong performance, it could serve as a backbone of various applications, e.g., multimodal dialogue systems \cite{zhu2023minigpt,li2023videochat,liu2023internchat}.
\end{itemize}

\section{Related Work}
\vspace{-0.2cm}
\paragraph{Multimodal Datasets.}
Vision-text data pairs are necessary to enable crossmodal learning. To learn vison-language representation effectively, these datasets should be large at scale and high at vision-text correlations. To this end, researches usually leverage existing web images with alt-text \cite{thomee2016yfcc100m,sharma2018conceptual,changpinyo2021conceptual,hu2022scaling,desai2021redcaps,he2023wanjuan,conghui2022opendatalab} and videos with ASR transcriptions \cite{miech2019howto100m,zellers2021merlot,zellers2022merlot,xue2022advancing,bain2021frozen,schuhmann2022laion,srinivasan2021wit} for scalable learning. With LAION-5B's introduction \cite{schuhmann2022laion}, researchers now have access to hundreds or millions or billions of image-text pairs, opening up new avenues for research on large-scale image-language pretraining.

For video-centric multimodal datasets, HowTo100M \cite{miech2019howto100m} collected instructional YouTube videos and exploited the corresponding ASR subtitles for learning joint representations. Zellers et al. \cite{zellers2021merlot,zellers2022merlot} and Xue et al. \cite{xue2022advancing} proposed YT-Temporal and HD-VILA for Audio-Visual-Language joint learning and high-resolution video crossmodal learning, respectively. On the other hand, Bain et al. \cite{bain2021frozen} found video-text alignment matters more than their quantities, so they produced WebVid \cite{bain2021frozen} where 10M videos with the corresponding alt-texts. This is frequently employed in recent video-language pretraining approaches \cite{umt}. Similarly, based on CC3M, Nagrani et al. proposed VideoCC3M \cite{nagrani2022learning} by transferring captions from image-text datasets to video ones. In this work, we target to present a large-scale video-language dataset with high-quality descriptions.

\vspace{-3mm}
\paragraph{Video Understanding.} 
Pretraining large-scale video-text models and fine-tuning them for downstream tasks has become the norm in the video-language field ~\cite{MiechASLSZ20,cpd,videoclip,umt,uniformerv2,videoclip,hu2022scaling,dou2022empirical,shen2021much,yao2021filip,videobert,actbert,wang2022internvideo,chen2022internvideo,zellers2021merlot,zellers2022merlot,zeng2023learning,zeng2023tvtsv2,chen2023videollm}. Early techniques~\cite{videobert,actbert} used pretrained visual and language encoders to obtain offline video and text features, but recent methods~\cite{cpd,MiechASLSZ20,hu2022scaling,dou2022empirical,videomae,wang2023videomae} highlight the advantages of end-to-end training. Common practices include two or three pretraining tasks, such as masked language modeling~\cite{lavender}, video-text matching~\cite{allinone}, video-text contrastive learning~\cite{videoclip,wang2022internvideo}, masked video modeling~\cite{videomae,wang2023videomae,wang2022internvideo}, and video-text masked modeling~\cite{violet}.

In the multimodal video context, VIOLET~\cite{violet} combined masked language and video modeling, while All-in-one~\cite{allinone} proposes a unified pretraining approach with a shared backbone, and LAVENDER~\cite{lavender} unified tasks through masked language modeling. Despite their success in multimodal benchmarks, these methods' reliance on limited video-text data hampers performance in video-only tasks like action recognition. Conversely, InternVideo \cite{wang2022internvideo} and UMT \cite{umt} combined masked modeling with crossmodal contrastive learning, leading to competitve performance in both video-only and video-language tasks. MERLOT Reserve~\cite{zellers2022merlot} exploited 20 million video-text-audio pairs for training joint video representations using contrastive matching, setting new standards in video recognition and visual commonsense reasoning. VALOR \cite{chen2023valor} also employed different modality encoders for video, audio, and text processing, and introduces video-to-text and audio-to-text pretasks to improve vision-audio-language learning.
To address modality entanglement in crossmodal learning, mPLUG-2 \cite{xu2023mplug} introduced a shared module across image, video, and text to encourage modality collaboration while reserving modality-specific modules for their differences. Similar to \cite{wang2022internvideo,uniformerv2}, VLAB \cite{he2023vlab} adapted a CLIP-pretrained ViT to model spatiotemporal variations and blends it with CLIP ViT with cross attention for handling both images and videos.

\begin{table*}[t]
% \scriptsize
\small
    \centering
    \resizebox{\textwidth}{!}{
    \begin{tabular}{r r r r r r r r r r} 
    \hline
    Dataset & Caption & Domain &\#Videos  & \#Clips & Len$_\text{Clip}$ & Len$_\text{Cap}$ & Dur(h) & Res \\\hline\hline
    MSR-VTT~\cite{xu2016msr} & Manual & open & 7.2K & 10K & 15.0 & 9.3 & 40 & 240P\\ 
    DideMo~\cite{anne2017localizing} & Manual & Flickr & 10.5K & 27K & 6.9 & 8.0 &87 & -\\ 
    LSMDC~\cite{rohrbach2017movie} & Manual & movie & 200 & 118K & 4.8 & 7.0 & 158 & 1080P\\
    YouCook2~\cite{zhou2018towards} & Manual & cooking & 2K & 14K  & 19.6 & 8.8 & 176 & -\\
    How2~\cite{sanabria2018how2} & Manual & instruct & 13.2K & 80K & 90.0 & 20.0 & 2K & -\\
    ANet Caption~\cite{krishna2017dense} & Manual & action & 20K & 100K & 36.0 & 13.5 & 849 & -\\
    %WebVid-2M~\cite{bain2021frozen} & open & & 2.5M & 18.0 & 12.0 & 13K & 360p\\
    VideoCC3M~\cite{nagrani2022learning} & Transfer & open & 6.3M & 10.3M & 10 & - & 17.5K & -\\
    WebVid10M~\cite{bain2021frozen} & Alt-text & open & 10.7M & 10.7M & 18.0 & 12.0 & 52K & 360P\\
    WTS70M~\cite{stroud2020learning} & Metadata & action & 70M & 70M & 10 & - & 194K & -\\
    HowTo100M~\cite{miech2019howto100m} & ASR & instruct & 1.2M & 136M & 3.6 & 4.0 & 134.5K & 240P\\
    HD-VILA-100M \cite{xue2022advancing} & ASR & open & 3.3M & 103M & 13.4 & 32.5 & 371.5K & 720P\\
    YT-Temporal-180M \cite{zellers2021merlot} & ASR & open & 6M & 180M & - & - & - & -\\
    \dataname~ (ours) & Generated & open & 7.1M & 234M & 11.7 & 17.6 & 760.3K & 720P*\\
    \hline
    \end{tabular}
    }
    \caption{Statistics of {\dataname} and its comparison with existing video-language datasets. *In {\dataname}, most videos (around 85\%) are in 720P and the remaining are in from 360P to 512P.}
    \label{tab:datasets}
    \vspace{-3mm}
\end{table*}

\section{\dataname: A Video-Centric Multimodal Dataset}
A high-quality video-text dataset at scale is a premise to conduct large-scale video-language learning and associated tasks. We identify three crucial factors in constructing this dataset: substantial temporal dynamics, rich and diverse semantics, and strong video-text correlations. To ensure high temporal dynamics, we gather videos retrieved using action/activity-based query words. For rich and varied semantics, we not only crawl trending videos across various categories but also deliberately increase the proportion of data consciously collected from various countries and languages.
To strengthen video-text correlations, we employ image captioning and language models to generate video descriptions from frame-specific annotations. Next, we elaborate the dataset construction process and discuss its statistics and characteristics.

%We illustrate the dataset holistically with statistics, then explain how it is built.

\subsection{Data Curation}
We collect videos from YouTube considering the diversity and richness of its data, and its support for academic usage. Totally we obtain 7 million public YouTube videos with an average duration of 6.4 minutes, 
covering 16 topics.
We ensure the uniqueness of our dataset by creating a database of YouTube video IDs and excluding any videos already present in publicly available datasets (released prior to April 2023). The data curation strategies are two-fold. On one hand, We select popular channels and the corresponding hot or high-rated videos from the categories e.g. news, gaming, etc., resulting in 2 million videos. On the other hand, we create a list of verbs related to actions/activities. With it, we also obtain 5.1 million videos by choosing the top retrieved ones.

\vspace{-2mm}
\paragraph{Defining Actions in Kinetics \& Motives for Queries.}
We define around 6.1K action phrases from American Time Use Survey (ATUS), public video datasets, and text corpus. Then they are refined both manually and automatically. We employ actions from ATUS from 2017 to 2022 \cite{caba2015activitynet}, merging them and removing the duplicates. 
For the referenced public video data, we leverage Kinetics \cite{k400}, SomethingSomething series \cite{goyal2017something,mahdisoltani2018effectiveness}, UCF101 \cite{soomro2012ucf101}, and so on. This provides us with 1103 action labels. Moreover, we access several visual grounding corpus \cite{cogrounding_2021_CVPR,yang2022tubedetr,li2017cvpr}. A language model \cite{brown2020gpt3} is employed to extract actions and their corresponding targets (if exist) to form phrases from the corpus, leading to 5001 actions with manual checking. Totally, we collect 6104 action queries for searching videos on YouTube.

\vspace{-2mm}
\paragraph{Collection Strategies.}
To ensure the quality of our dataset, we established specific crawling rules. We only collected videos that were between 10 seconds and 30 minutes in duration and had resolutions ranging from 360P to 720P. Videos with resolutions below 360P were excluded, and those above 720P were either downloaded in their 720P version or resized to 720P. In this process, we prioritize the highest available resolution.
To provide a comprehensive mutimodal dataset, we gather videos along with their audio, subtitles, titles, and summaries. Captions for the videos were generated automatically using a video captioning pipeline described in Section \ref{sec:caption}.

In formation, the collected multimodal data contain videos $\mathbf{V}$, their audios $\mathbf{A}$, metadata (title $\mathbf{W}^{\text{title}}$, video descriptions $\mathbf{W}^{\text{content}}$, query words $\mathbf{W}^{\text{query}}$, tags $\mathbf{W}^{\text{tag}}$, etc), subtitles (user generated contents or auto-generated ones), and more. Each video $\mathbf{V}$ could be treated as a sequence of clips $\{\mathbf{C}_i\}_{i=1,2,...}$, and we can segment their corresponding audio as $\{\mathbf{A}_i\}_{i=1,2,...}$ and ASR subtitles as $\{\mathbf{W}_i^{\text{asr}}\}_{i=1,2,...}$. For the metadata, we suppose clips share the same meta when they are sampled from the same video.

\vspace{-2mm}
\paragraph{Trimming.}
We segment videos (lasting an average of 5 minutes) into clips (for around 10 seconds) using scene variance. For starters, videos are cut into shorter ones based on their scene changes. We directly employ the corresponding filter in PySceneDetect \footnote{\url{https://github.com/Breakthrough/PySceneDetect}} with a threshold as 27. During this procedure, we also filter out clips in still or extreme dynamics (e.g. a browse of a photo gallery). After the filtering, we get total 234M video clips whose durations range from 2s to more than 30s.

\begin{figure*}[t]
    \centering
    \includegraphics[width=0.95\textwidth]{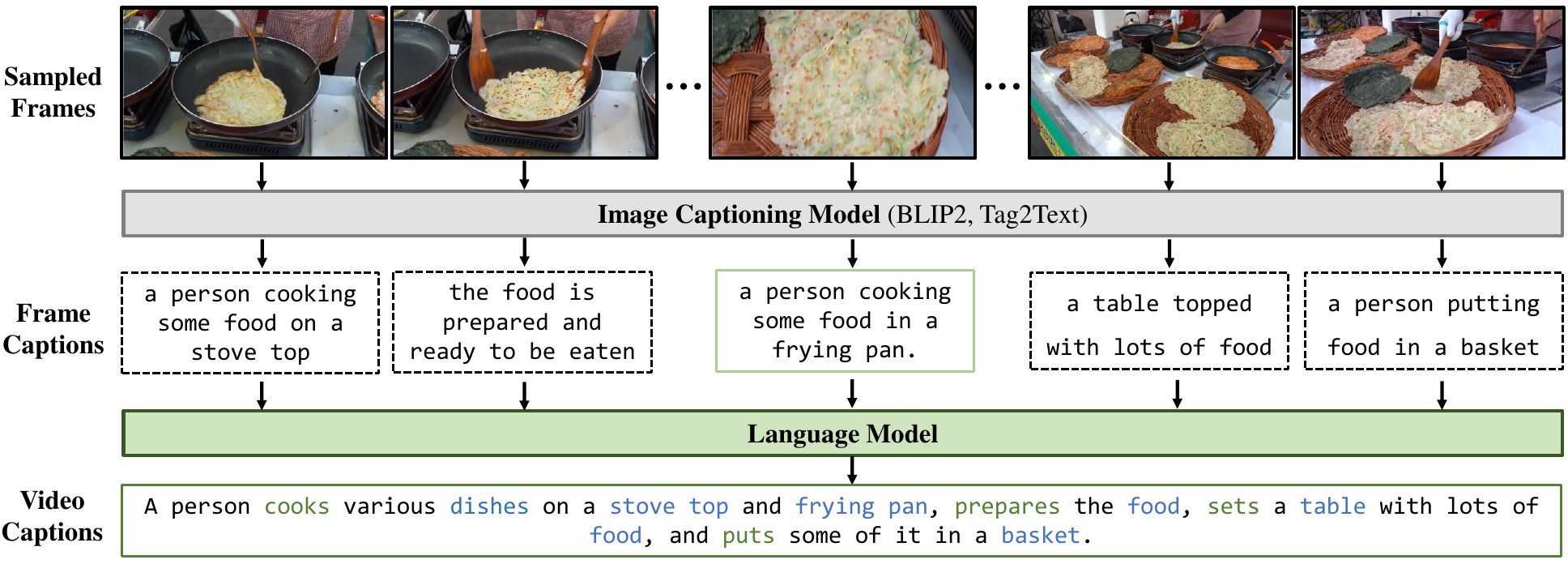}
    % This image is a place-holder. All texts are the same.
    \caption{The proposed multiscale video caption pipeline.
    The captions in coarse and fine scales are marked in \textcolor{green}{green} and \textcolor{darkGreen}{dark green}, respectively.
    }
    \label{fig:caption}
    \vspace{-0.3cm}
\end{figure*}

\subsection{Multiscale Video Captioning}
\label{sec:caption}
To generate video captions that are scalable, rich, and diverse, we employ a multiscale method with two distinct captioning strategies, as depicted in Figure \ref{fig:caption}.
On the finer scale, we simplify the video captioning process by concentrating on the common objects, actions, and scene descriptions within the video clip. We deliberately overlook intricate details such as subtle facial expressions \& movements, and other nuanced elements. On the coarser scale, we adopt the single-frame bias assumption from \cite{lei2022revealing} and exclusively caption the central frame of the video. Given our focus on brief clips (around 10 seconds) filtered via scene segmentation, most videos predominantly display consistent objects without substantial appearance alterations. This circumvents the identity-preserving issue when dealing with videos from image perspectives.
Technically, we employ the lightweight image captioning model Tag2Text \cite{huang2023tag2text} for the finer scale, which describes videos at low fps in a frame-by-frame manner. These individual image captions are then synthesized into a comprehensive video description using a pretrained language model \cite{raffel2020exploring,vicuna2023}. At the coarser scale, we use BLIP2 \cite{li2023blip} to caption the middle frame of the clip.

% Considering the presented video descriptions are 

\subsection{Statistics and Features}
We present the key statistics of {\dataname} with other popular video-language datasets in Table \ref{tab:datasets}. More detailed ones are given below.

\vspace{-2mm}
\paragraph{Diversity \& Richness.} 
% categories in videos / clips
% duration in videos / clips
% resolution in videos / clips
% languages in videos/ clips (regions)
% clipsim or xxxSIM
% concepts / noun / verb / ... stats
We collected videos from 16 popular categories with varying percentages, as illustrated in Figure \ref{fig:stat}. Unlike prior studies \cite{miech2019howto100m,xue2022advancing,zellers2021merlot}, we ensured diversity by selecting videos from countries with different languages instead of relying on a dominant language environment. The countries we sampled from include the UK, USA, Australia, Japan, Korea, China, Russia, and France, among others. In terms of duration, every video lasts 351.9s on average. Almost half (49\%) of the videos are five minutes or less, while a quarter (26\%) fall between five and ten minutes. Only 8\% of the videos are over 20 minutes long. Among the curated videos, 85\% were high-resolution (720P), while the remaining 15\% had lower resolutions ranging from 360P to 720P. Although the lower-resolution videos may not perform as well as the high-resolution ones in content generation tasks, they can still be useful in video-language representation learning, provided that they have appropriate captions.

{\dataname} exhibits diverse clip durations and caption lengths in the segmented clip level. The aesthetic scores and clip-caption similarities are distributed uniformly, as shown in Figure \ref{fig:clipstat}. The majority of clips are 0-10 seconds in length, accounting for 85\% of all clips (Figure \ref{fig:clipstat}: left). Approximately half of the clips have captions with 10-20 words, while one-third of the clip captions have fewer than 10 words. About 11\% of clips have long captions with more than 20 words.

We measured the aesthetic scores of all clips using an open-source model \cite{schuhmann2022laion}. We uniformly sampled four frames of each clip, calculated their aesthetic scores, and took the maximum score as the video aesthetic score. For clip-caption similarity computation, we used a video-language model called UMT \cite{umt}. We computed the cosine similarity between video embeddings and text embeddings, again using a uniform sampling of four frames for each clip.
Most clips score around 4-6 in terms of aesthetics, accounting for approximately 75\% of the data. For UMT-SIM, over 80\% of the clips scored between 0.3-0.4, with the remaining clips scoring around 0.2-0.3 or 0.4-0.5. Based on these computed aesthetics and UMT-SIM scores, we can generate different versions of {\dataname} to meet various requirements.

\iffalse
\begin{figure*}[t]
    \centering
    \includegraphics[width=0.9\textwidth]{fig/video_categories.png}
    % \vspace{-0.3cm}
    \caption{16 sampling scenarios in {\dataname}.
    }
    \label{fig:scenarios}
    \vspace{-0.3cm}
\end{figure*}
\fi

%\paragraph{Flexible Annotation}

\begin{figure*}[t]
    \centering
    \includegraphics[width=1.0\textwidth]{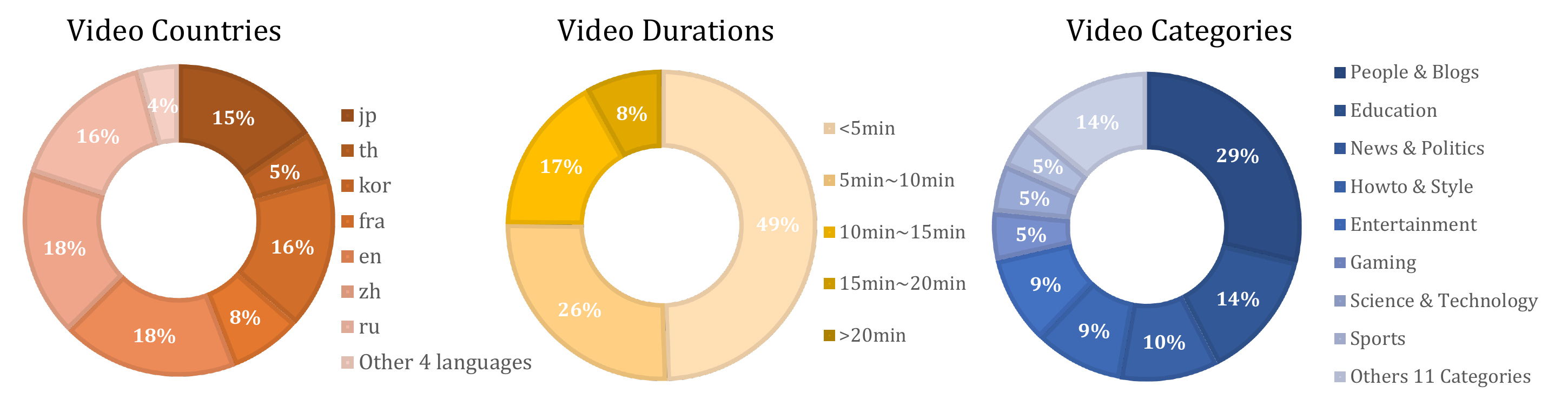}
    % \vspace{-0.3cm}
    \caption{Video statistics in {\dataname}.
It encompasses a diverse set of categories, gathered from multiple countries and averaging a duration of five minutes.
    }
    \label{fig:stat}
    \vspace{-0.3cm}
\end{figure*}

\begin{figure*}[t]
    \centering
    \includegraphics[width=1.0\textwidth]{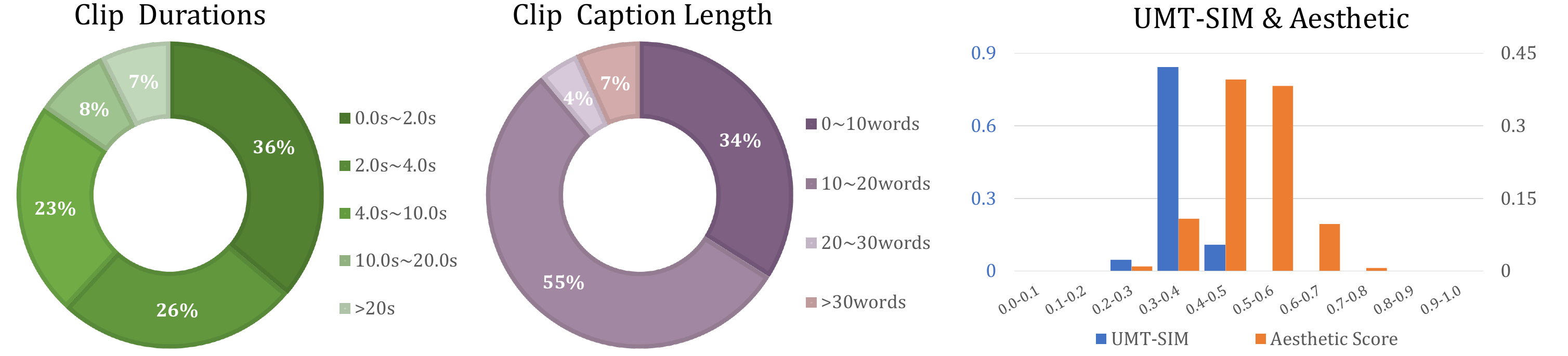}
    % \vspace{-0.3cm}
    \caption{Clip statistics in {\dataname}.  {\dataname} contains a diverse distribution of clip durations and caption lengths. It also offers aesthetic scores and multimodal similarity scores for each clip.
    }
    \label{fig:clipstat}
    \vspace{-0.3cm}
\end{figure*}

\begin{figure*}[t]
    \centering
    \includegraphics[width=0.9\textwidth]{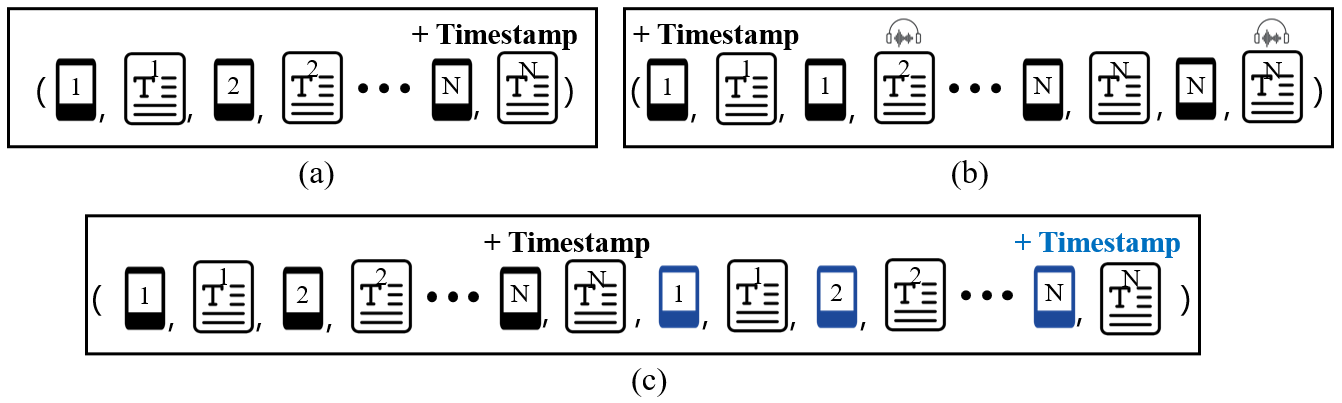}
    % \vspace{-0.3cm}
    \caption{Interleaved video-text data generation in {\dataname} with three formats.
    }
    \label{fig:interleaved}
    \vspace{-0.3cm}
\end{figure*}

\paragraph{Actionness.}
In terms of actionness, the {\dataname} dataset contains about ten times more verbs than the WebVid10M dataset. To evaluate this, we used the NLTK toolkit to analyze the number of verbs in captions, focusing on extracting and tagging all unique verbs. We found a total of 109,485 verbs in the WebVid10M caption dataset, while the {\dataname} dataset contained 212,155 unique instances of verbs. While these counts may not be entirely accurate due to our simple counting method, we believe they provide a rough indication of the actionness of the two datasets.

\subsection{Interleaved Video-Text Data Generation}
Utilizing the created video captions, we can develop an integrated video-text dataset for in-context video learning, allowing video-based sequence models to perform new tasks without additional training. Previous research, such as Flamingo \cite{flamingo,openflamingo}, Kosmos-1 \cite{kosmos}, and Multimodal C4 \cite{zhu2023multimodal}, confirms that pretraining on the interleaved image-text sequences results in significant multimodal in-context abilities. To the best of our knowledge, a large-scale interleaved video-text dataset has not yet been established. Our work represents the initial step in creating and making it publicly available.

We create {\dataname}-ICL, containing 7.1M interleaved video-text data pairs. We propose three distinct methods for organizing clips and their captions:

 $\bullet$ Arrange clips and their descriptions sequentially based on their temporal order within the same video, as illustrated in Figure \ref{fig:interleaved} (a).

 $\bullet$ Enhance diversity in interleaved video-text items by assigning ASR text to a used clip in addition to its caption, as demonstrated in Figure \ref{fig:interleaved} (b).
 
 $\bullet$ Extend method 1 by concatenating two interleaved multimodal items, creating a video-centric dialogue simulating user queries involving multiple videos (Figure \ref{fig:interleaved} (c)).

\begin{table*}[h!]\centering
    \begin{minipage}{0.99\columnwidth}\vspace{0mm}    
    \centering
    \begin{tcolorbox} 
        \centering
        \hspace{-6mm}
        \begin{tabular}{p{0.99\columnwidth}}
        \hspace{1mm}
        \begin{minipage}{0.99\columnwidth}

        \texttt{[..., "the inside of a home has a rug and a light on.", \textcolor{gray}{"♪ We could leave the Christmas lights up til January ♪"}, ..., "woman with blond hair playing guitar", \textcolor{gray}{"♪ Have I known you 20 seconds or 20 years? ♪",}}
        {\makebox[0pt][l]{\hspace{0pt}\raisebox{-0.3ex}{{\includegraphics[height=25pt]{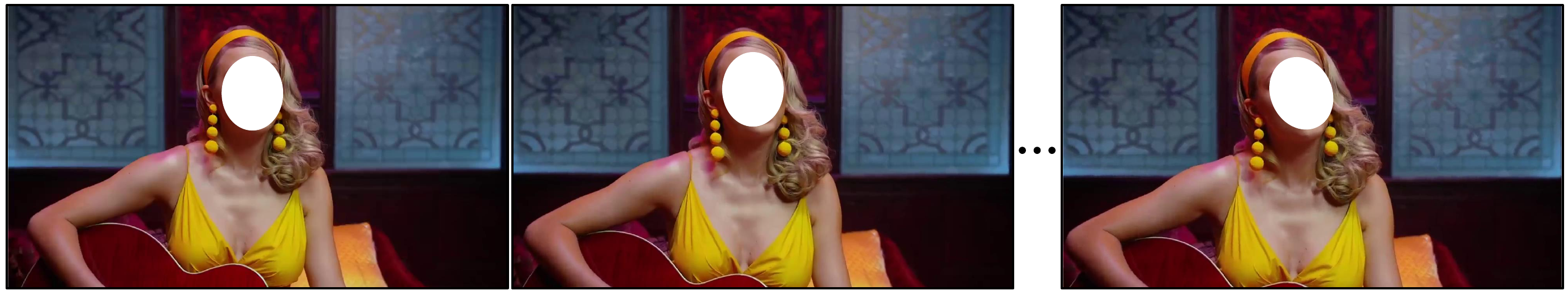}}}}}
        \texttt{~~~~~~~~~~~~~~~~~~~~~~~~~~~~, "close-up of a bathroom sink with soap bubbles and other items", "a bathroom is seen with a sink and two lights", "a woman swiming inside of a fishbowl with a ladder and a man", \textcolor{gray}{"♪ Can I go wher you go? ♪",}}
        {\makebox[0pt][l]{\hspace{0pt}\raisebox{-0.3ex}{{\includegraphics[height=25pt]{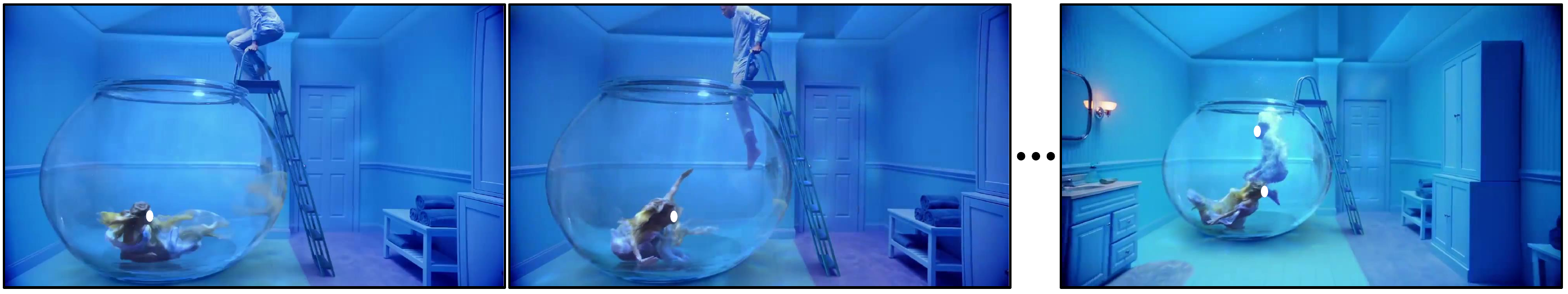}}}}}
        
        \texttt{, "devils roll the dice, angels roll their eyes",\textcolor{gray}{"♪ And, take me out, and take me home ♪" ,}..., "the man is standing in a room with pink carpet",\textcolor{gray}{"♪ You're my, my ♪"}, "a woman in yellow is dancing with a man in a red room", \textcolor{gray}{"♪ My, My lover ♪",} }\\       
        {\makebox[0pt][l]{\hspace{0pt}\raisebox{-0.3ex}{{\includegraphics[height=25pt]{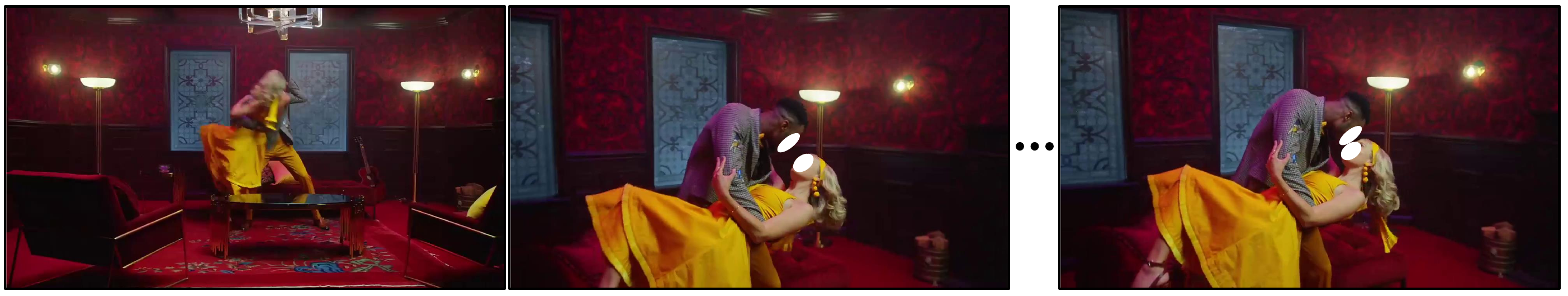}}}}}
        \texttt{~~~~~~~~~~~~~~~~~~~~~~~~~, "a woman is sitting on a chair, playing a guitar and a woman holding a balloon", \textcolor{gray}{"♪ ♪ ♪",} "two men smiling while holding wine glasses and drinking beer", \textcolor{gray}{"♪ We could let our friends crash in the living room ♪"} ...]}
        \end{minipage}
        \end{tabular}
    \end{tcolorbox}
    \vspace{-2mm}
    \caption{\textbf{Interleaved video-text data format (b) in {\dataname}.} The caption and ASR transcript of each clip is shown in black and gray, respectively. We can achieve interleaved video-text data format (a) by abandoning ASR transcripts. To obtain data format (c), we concatenate multiple videos with interleaved video-text data (a).
}
    \label{tab:video-icl1}
\end{minipage}
    \vspace{-2mm}
\end{table*}
 
One visual example of these arrangements is provided in Table \ref{tab:video-icl1}.

\begin{figure*}[t]
    \centering
    \includegraphics[width=0.7\textwidth]{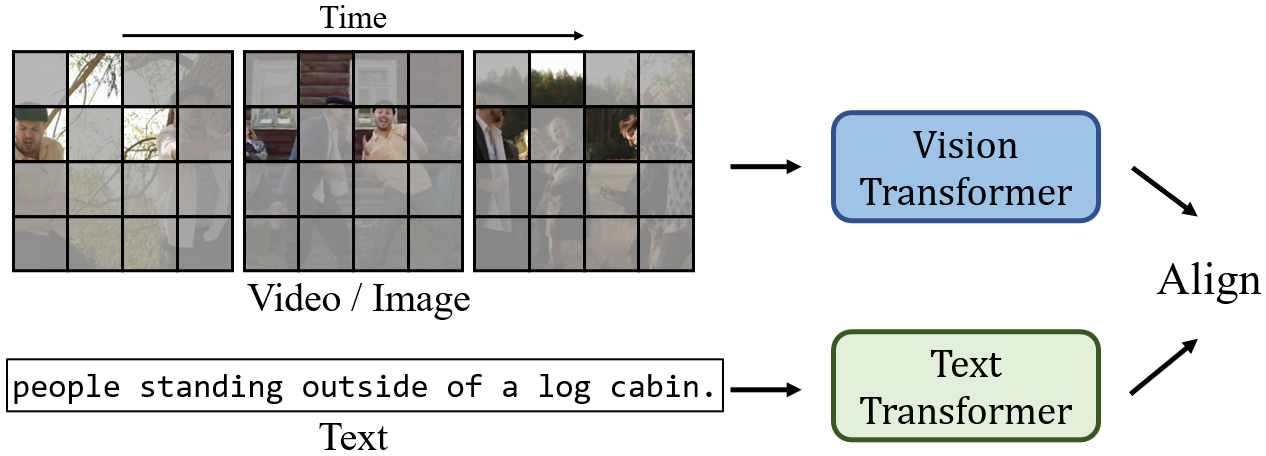}
    \caption{Framework of {\modelname}.
    }
    \label{fig:frame}
    \vspace{-0.3cm}
\end{figure*}
\section{{\modelname: Learning Video-Text Representation at Scale}}
Built upon CLIP \cite{radford2021learning}, we make a simple video-text pretraining baseline {\modelname}. It consists of a video encoder (ViT) \cite{dosovitskiy2021image} and a text encoder, as given in Figure \ref{fig:frame}. Both modules are initialized from the corresponding CLIP components. We update the native attention in the video encoder to spatiotemporal attention while maintaining other design elements. For efficient learning, we apply masking to videos in pre-training. The optimization target is the contrastive loss between input video and text embeddings.

\vspace{-2mm}
\paragraph{Video \& Text Encoders with Masking Learning.}
Our video encoder uses a standard ViT with spatiotemporal attention. We apply random patch masking following MAE-based methods \cite{videomae,wang2023videomae} to the input videos. It significantly alleviates the computational burden. The used text encoder is also a transformer followed by \cite{radford2021learning,schuhmann2022laion}.

\vspace{-2mm}
\paragraph{Unmasked Video-Text Pretraining.} We feed all visual tokens into the video transformer instead of just the masked ones towards the end of the pretraining process. This helps bridge the gap between pretraining and downstream applications where the full video is used as input. We perform unmasked training for 0.5 epochs with a learning rate of 4e-6.

\textbf{Training Objectives.}
Our framework optimizes video-text alignment. It minimizes InfoNCE loss \cite{oord2018representation} using global video and text features, as
\begin{equation} \small
    \mathcal{L}_{\text{C}} = \mathcal{L}_{\text{C}}^{\mathbf{V} \rightarrow \mathbf{T}}+\mathcal{L}_{\text{C}}^{\mathbf{T} \rightarrow \mathbf{V}} = - \sum^N_{i=1}{\text{log} \frac{\text{exp}(\text{sim}(f^{\mathbf{V}}_i, f^{\mathbf{T}}_i) / \tau)}{\sum^N_{j=1}{\text{exp}(\text{sim}(f^{\mathbf{V}}_i, f^{\mathbf{T}}_j) / \tau)}}} - \sum^N_{i=1}{\text{log} \frac{\text{exp}(\text{sim}(f^{\mathbf{T}}_i, f^{\mathbf{V}}_i) / \tau)}{\sum^N_{j=1}{\text{exp}(\text{sim}(f^{\mathbf{T}}_i, f^{\mathbf{V}}_j) / \tau)}}},
\end{equation}
where $f^{\mathbf{V}}$ and $f^{\mathbf{T}}$ denote the learned video and text embeddings, respectively. $\text{sim}(\cdot)$ computes the cosine similarity between two features. $\tau$ is the learnable temperature.

\vspace{-2mm}
\paragraph{Implementation.} {\modelname} is learned with 64 NVIDIA A100 GPUs for 3 days with 50M video-text pairs. We introduce DeepSpeed and FlashAttention \cite{dao2022flashattention} for training and inference acceleration.

We learn {\modelname} on five subsets of {\dataname} and evaluated its performance on popular video-related benchmarks using full-finetuned and zero-shot settings. We sample subsets {\dataname}-10M, {\dataname}-50M, and  {\dataname}-200M randomly. 
For {\dataname}-10M-DIV, 
%we hypothesize that the diversity of the training video clips matters more than its quantity. To construct {\dataname}-10M-DIV, 
we prioritize to sample clips from different videos first, then we sample clips with varying probabilities according to the video length where they are extracted. The longer their source video is, the lower chance they are sampled. For {\dataname}-10M-FLT, we employ the sampling strategy of {\dataname}-10M-DIV and select clips with UMT-SIM scores ranking among the top 30\% to ensure high quality.
%We compare {\dataname}-10M and {\dataname}-10M-DIV / -FLT with WebVid10M, while we use {\dataname}-50M and {\dataname}-200M to further validate the data scalability of video-language contrastive learning.

\vspace{-1mm}
\subsection{Transferable Video Representation Performance} 
\vspace{-1mm}
\textbf{Action Recognition.} \label{sec:zs-action}
In addition to OpenAI's CLIP-L (CLIP400M~\citep{radford2021learning}) and LAION (DataComp-1B~\citep{gadre2023datacomp}), we also include EVA-CLIP-L/14 and EVA-CLIP-E/14~\citep{sun2023eva} for comparison. More experimental settings are given in App. \ref{sec:vclip_set}.
%EVA-CLIP~\citep{sun2023eva} uses a distillation model with masked modeling to learn image-text joint representation.

\begin{table}[t]
% \scriptsize
\centering
% \resizebox{\textwidth}{!}{
% \scriptsize
\small
\setlength{\tabcolsep}{1.0 mm}{
  \begin{tabular}{llllllll}
    \toprule
     \multirow{2}{*}{Method}& \multirow{2}{*}{Training Data} &\multicolumn{2}{c}{K400} & \multicolumn{2}{c}{K600} & \multicolumn{2}{c}{K700}\\
     & & top-1 ($\uparrow$) & AVG ($\uparrow$) & top-1 ($\uparrow$) & AVG ($\uparrow$) &top-1 ($\uparrow$) & AVG ($\uparrow$)\\
    \midrule
     CLIP & CLIP400M & 58.42 & 70.14 & 55.11 & 67.16 & 46.12 & 58.38 \\
     CLIP & DataComp-1B & 56.14 & 67.67 & 54.15 & 65.83 & 45.36 & 57.01 \\
     EVA-CLIP-L & Merged-2B & - & 65.00 & - & 64.90 & - & 59.10 \\
     EVA-CLIP-E & LAION-2B & - & 69.80 & - & 69.30 & - & 63.40 \\
     \modelname & +WebVid10M & 59.88 & 71.03 & 58.66 & 69.84 & 50.23 & 61.86\\
     \hline
     \modelname & +{\dataname}-10M & 56.68 & 68.17 & 54.67 & 66.28 & 46.53 & 58.73\\
     \modelname & +{\dataname}-50M & 57.18 & 68.93 & 55.36 & 67.07 & 47.00 & 59.36\\
     \modelname & +{\dataname}-200M & 59.80 & 71.09 & 57.80 & 69.34 & 49.30 & 61.25\\
     \modelname & +{\dataname}-10M-DIV & 63.00 & 74.15 & 60.68 & 72.07 & 52.50 & 64.59\\
    \modelname & +{\dataname}-10M-FLT & \textbf{64.80} & \textbf{75.70} & \textbf{62.20} & \textbf{73.53} & \textbf{54.30} & \textbf{66.38}\\
     
    \bottomrule
  \end{tabular}
}
% }
% \vspace{-3mm}
\caption{Zero-shot action recognition results on Kinetics 400/600/700.}
\vspace{-3mm}
\label{tab:zs_recognition}
\end{table}

\begin{table}[t]
% \scriptsize
\centering
% \resizebox{\textwidth}{!}{
% \scriptsize
\small
\setlength{\tabcolsep}{3.0 mm}{
  \begin{tabular}{llllll}
    \toprule
     \multirow{2}{*}{Method}& \multirow{2}{*}{Training Data} &\multicolumn{2}{c}{K400} & \multicolumn{2}{c}{SthSthV2} \\
     & & top-1 ($\uparrow$) & top-5 ($\uparrow$) & top-1 ($\uparrow$) & top-5 ($\uparrow$)\\
    \midrule
     CLIP & CLIP400M & 86.7 & 97.2 & 70.1 & 92.5 \\
     CLIP & DataComp-1B & 85.6 & 96.8 & 68.9 & 91.8\\
     % EVA-CLIP-L~\citep{sun2023eva}& Merged-2B & - & 65.00 & - & 64.90 & - & 59.10 \\
     % EVA-CLIP-E~\citep{sun2023eva} & LAION-2B & - & 69.80 & - & 69.30 & - & 63.40 \\
     \modelname & +WebVid10M & 85.0 & 96.8 & 68.7 & 91.9\\
     \hline
     % \modelname & +{\dataname}-10M & 56.68 & 68.17 & 54.67 & 66.28 & 46.53 & 58.73\\
     % \modelname & +{\dataname}-50M & 57.18 & 68.93 & 55.36 & 67.07 & 47.00 & 59.36\\
     \modelname & +{\dataname}-10M-FLT & 86.8 & 97.5 & 71.2 & 93.2 \\
     \modelname & +{\dataname}-10M-FLT+K710 & 88.0 & 97.8 & 71.8 & 93.6 \\
     \modelname & +{\dataname}-200M & 87.9 & 97.9 & 73.6 & 94.9 \\
     \modelname & +{\dataname}-200M+K710 & \textbf{88.7} & \textbf{98.2} & \textbf{74.2} & \textbf{95.0} \\
     % \modelname & +{\dataname}-10M-DIV & 63.00 & 74.15 & 60.68 & 72.07 & 52.50 & 64.59\\
    % \modelname & +{\dataname}-10M-FLT & \textbf{64.80} & \textbf{75.70} & \textbf{62.20} & \textbf{73.53} & \textbf{54.30} & \textbf{66.38}\\
     
    \bottomrule
  \end{tabular}
}
% }
% \vspace{-3mm}
\caption{Fine-tuned action recognition results on Kinetics 400 and SomethingSomethingV2.}
% \vspace{-5mm}
\label{tab:ft_recognition}
\end{table}

\vspace{-1mm}
\textit{Zero-Shot.} 
%Table \ref{tab:zs_recognition} shows that when using WebVid10M, {\modelname} surpasses OpenAI's CLIP~\citep{radford2021learning}, EVA-CLIP-L, and EVA-CLIP-E on nearly all datasets. The exception is K700 where EVA-CLIP-E presents superior mean values for top-1 and top-5, indicating that model capacity may compensate for the data gap in some cases.
Table \ref{tab:zs_recognition} shows that when trained on \dataname-10M-FLT, {\modelname} outperforms all other methods, including EVA-CLIP-E. This result validates {\dataname}'s effectiveness in learning video-text embeddings. Note that {\modelname} with \dataname-10M-FLT sets new records on zero-shot action recognition in Kinetics 400/600/700, demonstrating a significant performance boost compared to {\modelname} with WebVid10M or other models.
Moreover, {\modelname} trained on \dataname-10M-FLT exceeds its performance on \dataname-200M. %This outcome suggests that scaling video-text data can enhance representation learning. However, data quality appears more critical than data scale in these evaluations.
Normally, we would expect the model trained on \dataname-200M to perform better than those on -10M-DIV or -FLT, given that the latter two subsets derive from the former. Unless this discrepancy results from improper learning, we conjecture that false negative samples could severely impede video-text contrastive learning if we don't purposefully reduce the number of clips taken from the same video.
Specifically, we hypothesize that clips from the same video share similar representations and captions. Contrastive learning, however, assumes these clips to be different. This situation also undermines the significance of using a large batch size in current training since it increases the probability of encountering more false negatives. We believe this assumption is applicable to other video tasks as well and plan to explore this further in the future.

\begin{table}[t]
\centering
\small
% \scriptsize
\resizebox{\textwidth}{!}{
\setlength{\tabcolsep}{1.5 mm}{
  \begin{tabular}{lcllllllllll}
    \toprule
    \multirow{2}{*}{Method}& \multirow{2}{*}{Data}& \multicolumn{2}{c}{MSR-VTT} & \multicolumn{2}{c}{LSMDC} & \multicolumn{2}{c}{DiDeMo} & \multicolumn{2}{c}{MSVD} & \multicolumn{2}{c}{ANet} \\
     & & T2V & V2T & T2V & V2T & T2V & V2T & T2V & V2T & T2V & V2T \\
    \midrule
     CLIP~ & CLIP400M & 29.0 & 25.8 & 13.9 & 15.2 & 11.5 & 19.1 & 37.9 & 60.0 & 8.3 & 12.2\\
     CLIP~ & DataComp-1B & 30.4 & 24.2 & 13.9 & 11.9 & 12.7 & 18.7 & 40.5 & 57.2 & 9.1 & 13.2\\
     CLIP4Clip \citep{clip4clip} & +HowTo100M & 32.0 & - & 15.1 & - & - & - & 38.5 & - & - & -\\
     \modelname~ & +WebVid10M & 35.6 & 33.1 & 16.5 & 13.4 & 14.5 & 23.3 & 45.3 & 69.0 & 12.4 & 19.0\\
     \midrule
     \modelname~ & +{\dataname}-10M & 36.4 & 37.1 & 17.1 & 15.0 & 16.4 & 25.9 & 45.2 & 69.8 & 13.5 & 23.4\\
     \modelname~ & +{\dataname}-50M & 39.7 & 40.7 & 18.0 & 16.7 & 16.7 & 26.4 & 46.5 & 72.2 & 13.6 & 23.2\\
     \modelname~ & +{\dataname}-200M & 39.3 & 39.5 & 18.3 & 16.6 & 17.1 & 25.5 & 47.3 & 70.0 & 13.7 & 21.6\\
     \modelname~ & +{\dataname}-10M-DIV & 41.5 & \textbf{41.6} & 18.5 & \textbf{17.4} & 17.7 & 26.2 & 48.6 & 71.9 & 14.8 & 23.4\\
     \modelname~ & +{\dataname}-10M-FLT & \textbf{42.4} & 41.3 & \textbf{20.1} & 16.9 & \textbf{18.4} & \textbf{27.9} & \textbf{49.1} & \textbf{75.1} & \textbf{15.1} & \textbf{24.0}\\
    \bottomrule
  \end{tabular}
}
}
% \vspace{-4mm}
\caption{Results of zero-shot video retrieval on MSR-VTT, LSMDC, DiDeMo, MSVD, and ANet.}
% \vspace{-4mm}
\label{tab:zs_retrieval}
\end{table}

\begin{table}[t]
\centering
\small
% \scriptsize
\resizebox{\textwidth}{!}{
\setlength{\tabcolsep}{1.5 mm}{
  \begin{tabular}{lcllllllllll}
    \toprule
    \multirow{2}{*}{Method}& \multirow{2}{*}{Data}& \multicolumn{2}{c}{MSR-VTT} & \multicolumn{2}{c}{LSMDC} & \multicolumn{2}{c}{DiDeMo} & \multicolumn{2}{c}{MSVD} & \multicolumn{2}{c}{ANet} \\
     & & T2V & V2T & T2V & V2T & T2V & V2T & T2V & V2T & T2V & V2T \\
    \midrule
    CLIP & CLIP400M & 38.2 & 38.7 & 22.5 & 22.6 & 32.2 & 33.9 & 67.3 & 69.9 & 26.1 & 26.9\\
    CLIP & DataComp-1B & 37.2 & 37.5 &  18.7 & 18.5 & 33.5 & 34.2 & 66.3 & 70.2 & 24.5 & 25.8\\
     CLIP4Clip \citep{clip4clip} & +HowTo100M & 45.6 & 45.9 & 24.3 & 23.8 & 43.0 & 43.6 & 45.2 & 48.4 & 40.3 & 41.6\\
     \modelname~ & +WebVid10M & 50.8 & 49.3 & 27.3 & 28.4 & 48.1 & 48.5 & 76.7 & \textbf{81.2} & 44.5 & 43.2\\
     \midrule
     \modelname~ & +{\dataname}-10M & 51.8 & 49.7 & 28.5 & 29.4 & 49.5 & 50.6 & 77.2 & \textbf{80.0} & 49.7 & 48.4\\
     \modelname~ & +{\dataname}-50M & 52.8 & 52.2 & 30.9 & 30.9 & 49.4 & 48.7 & 78.1 & \textbf{80.0} & 49.7 & 49.0\\
     \modelname~ & +{\dataname}-200M & 53.7 & \textbf{53.4} & 29.3 & 31.3 & 51.1 & 50.8 & \textbf{79.9} & 78.4 & \textbf{52.8} & \textbf{51.1}\\
     \modelname~ & +{\dataname}-10M-DIV & \textbf{55.0} & 53.3 & 32.0 & 30.0 & \textbf{51.7} & \textbf{52.1} & 75.8 & 77.8 & 50.4 & 48.9\\
     \modelname~ & +{\dataname}-10M-FLT & 52.5 & 51.8 & \textbf{33.0} & \textbf{32.5} & 49.4 & 50.2 & 77.2 & 79.0 & 49.8 & 48.1\\
    \bottomrule
  \end{tabular}
}}
% \vspace{-4mm}
\caption{Results of fine-tuned video retrieval on MSR-VTT, LSMDC, DiDeMo, MSVD, and ANet.}
% \vspace{-6mm}
\label{tab:retrieval}
\end{table}

\vspace{-1mm}
\textit{Fine-tuned.} In Table \ref{tab:ft_recognition}, note when comparing {\modelname} trained on {\dataname} with image CLIP models or {\modelname} trained with WebVid, there is a clear increase in accuracy. Unlike the zero-shot results, when {\modelname} is pretrained with a larger number (200M) of video-text data pairs, it achieves higher accuracy in fine-tuned recognition tasks (87.9\% in K400 and 73.6\% in SthSthV2) compared to when pretrained (86.8\% in K400 and 71.2\% in SthSthV2) with fewer data (10M). This suggests that {\dataname} provides greater benefits for fine-tuned action-related tasks. The decrease in performance of {\modelname} with WebVid highlights the importance of addressing the distribution gap between WebVid and the action videos used for evaluation, emphasizing the need to collect videos with evident temporal dynamics.

\vspace{-1mm}
\textbf{Video-Text Retrieval.}
We evaluate the video retrieval performance of baselines and {\modelname} using different pretraining datasets on five popular benchmarks~\citep{caba2015activitynet,xu2016msr,lsmdc, anne2017localizing, msvd}, as shown in Table \ref{tab:zs_retrieval} and \ref{tab:retrieval}. We uniformly sample eight frames from the input videos. For the CLIP models from OpenAI \citep{radford2021learning} and LAION \citep{schuhmann2022laion}, we utilize their officially released ViT-L models and extract video embeddings by averaging the computed frame-wise image embeddings. Our {\modelname} directly predicts video embeddings. For evaluating retrieval performance, \textit{we report R@1 scores for both text-to-video (t2v) and video-to-text (v2t) tasks} in \ref{tab:zs_retrieval} and \ref{tab:retrieval}.

\vspace{-1mm}
Both Table \ref{tab:zs_retrieval} and \ref{tab:retrieval} demonstrate that video-language pretraining is crucial for enhancing fine-tuned and zero-shot retrieval performance. This point is substantiated by the comparison between CLIP and {\modelname} using \dataname-50M. Table \ref{tab:zs_retrieval} exhibits a boost of nearly 4-10 points across different benchmarks in the zero-shot setting. Meanwhile, Table \ref{tab:retrieval} shows an increase of approximately 10 points across all R@1 scores in the fine-tuned setting.

\vspace{-1mm}
\textit{Zero-Shot.} 
Table \ref{tab:zs_retrieval} reveals \dataname-10M outperforms WebVid when employing the same method, {\modelname}, with an average increase of 6.3\% in R@1 across nearly all benchmarks. This improvement can be further amplified by diversifying the training clips used, as \dataname-10M-DIV and -FLT surpass WebVid on {\modelname} with gains in R@1 of 14.0\% and 17.1\%, respectively.
These results underline, once again, the effectiveness of the correspondence between our generated video captions and their corresponding videos. 
% However, in MSVD, {\modelname} using \dataname-10M displays a slightly lower R@1 score in text-to-video retrieval compared to when using WebVid10M (45.2 vs. 45.3). We believe this slight discrepancy does not undermine our assertion regarding the efficacy of {\dataname}.
Comparing CLIP4Clip using HowTo100M with {\modelname} using WebVid10M or \dataname-10M shows that the correlation between video and text influences performance more significantly than their quantity.
Moreover, the zero-shot performance demonstrates that the video-text representation learned using {\dataname} is transferable. This claim is supported by its superior performance across multiple video retrieval benchmarks.

\vspace{-1mm}
\textit{Fine-Tuned.} 
Table \ref{tab:retrieval} exhibits a noticeable improvement when transitioning from \dataname-10M to WebVid10M while using {\modelname} for both t2v and v2t retrieval across almost all datasets. On average, there is a 3.7\% increase in t2v R@1 across all benchmarks, with particularly significant rise observed in ActivityNet (an increase of over 11.9\%).
However, {\modelname} using WebVid10M yields better v2t R@1 scores than when using \dataname-10M (81.2 vs. 80.0). We believe this does not alter the overall trend that \dataname-10M generally provides more advantage to {\modelname} than WebVid10M does.

\vspace{-1mm}
The benefits of used video data become even more apparent when comparing \dataname-10M-DIV or \dataname-10M-FLT with WebVid10M. Their overall increases are 5.8\% and 5.1\%, respectively. Despite these improvements, issues related to data diversity persist.

\begin{figure}[t]
\centering
% \vspace{-9mm}
\begin{minipage}[t]{0.485\textwidth} 
\centering
\includegraphics[width=1\textwidth]{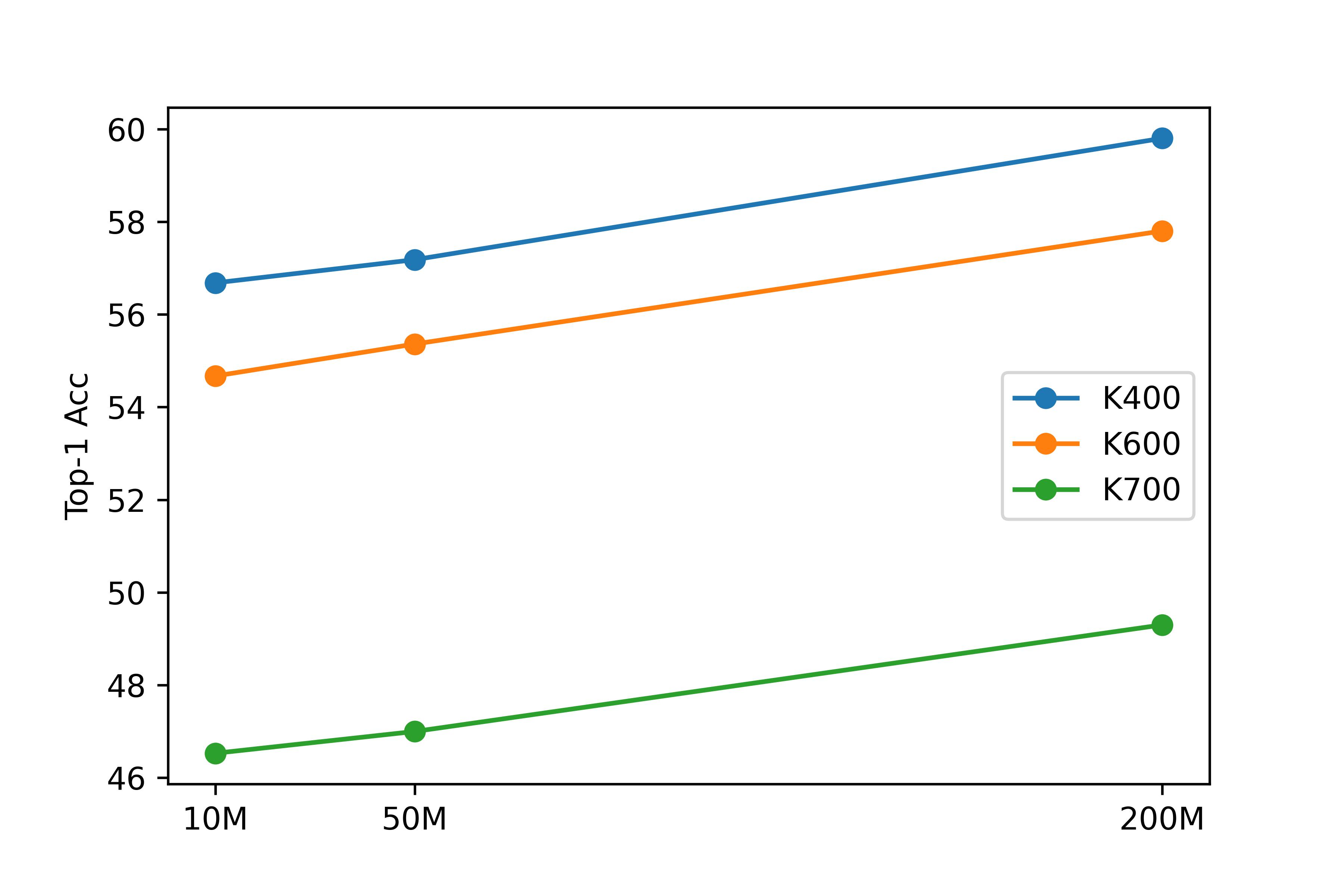}
% \vspace{-9mm}
\caption{Zero-shot action recognition (top-1 accuracy) on Kinetics-400 / -600 / -700.}
\end{minipage}
% \vspace{-2mm}
\begin{minipage}[t]{0.485\textwidth} 
\centering
\includegraphics[width=1\textwidth]{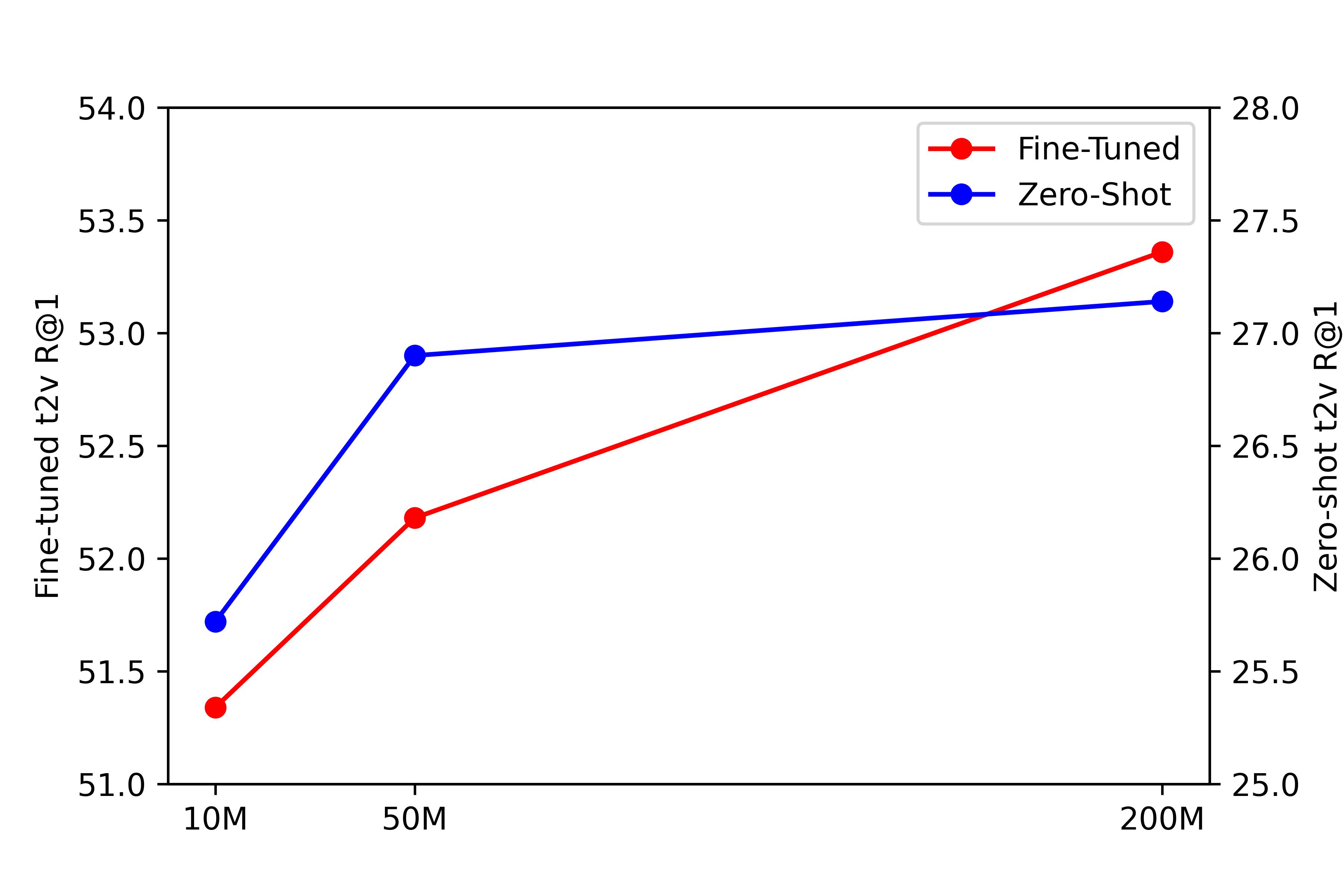}
% \vspace{-9mm}
\caption{Video retrieval average performance (text-to-video R@1) across five datasets.}
\end{minipage}
% \vspace{-2mm}
\end{figure}

\vspace{-1mm}
\textbf{Data Scaling and Issues.} 
Figure 7 and 8 illustrate how {\modelname}'s performance changes in zero-shot and fine-tuning settings when varying the scale of {\dataname}. In both scenarios, increasing the data scale results in significant increases in performance.
As shown in Figure 7, \modelname's discriminative ability linearly increases with the increasing volume of training videos used (10M $\rightarrow$ 200M). Meanwhile, Figure 8 shows that the retrieval performance increase becomes marginal when scaling the training data beyond 50M.
It's vital to note our model is trained using only contrastive loss without employing popular designs such as matching head and its corresponding loss. Consequently, this retrieval result doesn't allow for any definitive conclusions about whether there exists a turning point after which scaling up the training videos becomes less beneficial currently. More explorations are necessary in these retrieval experiments.
However, these findings generally suggest that enhancing the scale of pretraining data can improve the transferability of the learned representation.

% Moreover, Table \ref{tab:zs_recognition}, \ref{tab:zs_retrieval}, and \ref{tab:retrieval} demonstrate that {\modelname} trained with \dataname-10M-FLT or -DIV generally yields better action recognition and retrieval performance than the model trained on {\dataname} using a naive sampling strategy.
% As discussed in Section \ref{sec:zs-action}, we attribute this result to false negatives arising from sampling clips from the same video within a training batch. Implementing an intuitive sampling strategy during training to alleviate this issue may further enhance model performance by improving data exploitation. We leave this aspect for future exploration.

\begin{table}[t]
    \vspace{-3mm}
    \centering
    \setlength{\tabcolsep}{1.0 mm}{
    % \small
    \scriptsize
    \begin{tabular}[htop]{c c l l l l}
        \toprule
        \multirow{2}{*}{Method}& \multirow{2}{*}{Training Data} &\multicolumn{3}{c}{UCF-101} & \multicolumn{1}{c}{MSR-VTT}\\
     & & IS ($\uparrow$) & FID ($\downarrow$) & FVD ($\downarrow$)  & CLIPSIM ($\uparrow$)\\
        \midrule VideoCrafter\footnote{\url{https://github.com/VideoCrafter/VideoCrafter}} & WebVid10M & 18.26 & 66.95 & 910.87 & 0.2875 \\
        VideoFusion \footnote{\url{https://huggingface.co/spaces/damo-vilab/modelscope-text-to-video-synthesis}} & WebVid10M & 17.49 & 75.77 & 639.90 & 0.2795 \\
        \midrule
        t2v baseline & WebVid10M & 13.97 & 98.25 & 705.25 & 0.2657 \\
        t2v baseline & WebVid10M+{\dataname}18M & \textbf{21.04}$_{+7.07}$ & \textbf{60.25}$_{-38.00}$ & \textbf{616.51}$_{-88.74}$ & \textbf{0.2951}$_{+0.0294}$ \\
        \bottomrule
    \end{tabular}
    }
    % \vspace{-3mm}
    \caption{{Zero-shot text-to-video generation performance.}
    }
    \label{tab:video_gen}
    % \vspace{-6mm}
\end{table}

\begin{figure*}[t]
    \centering
    \includegraphics[width=0.9\textwidth]{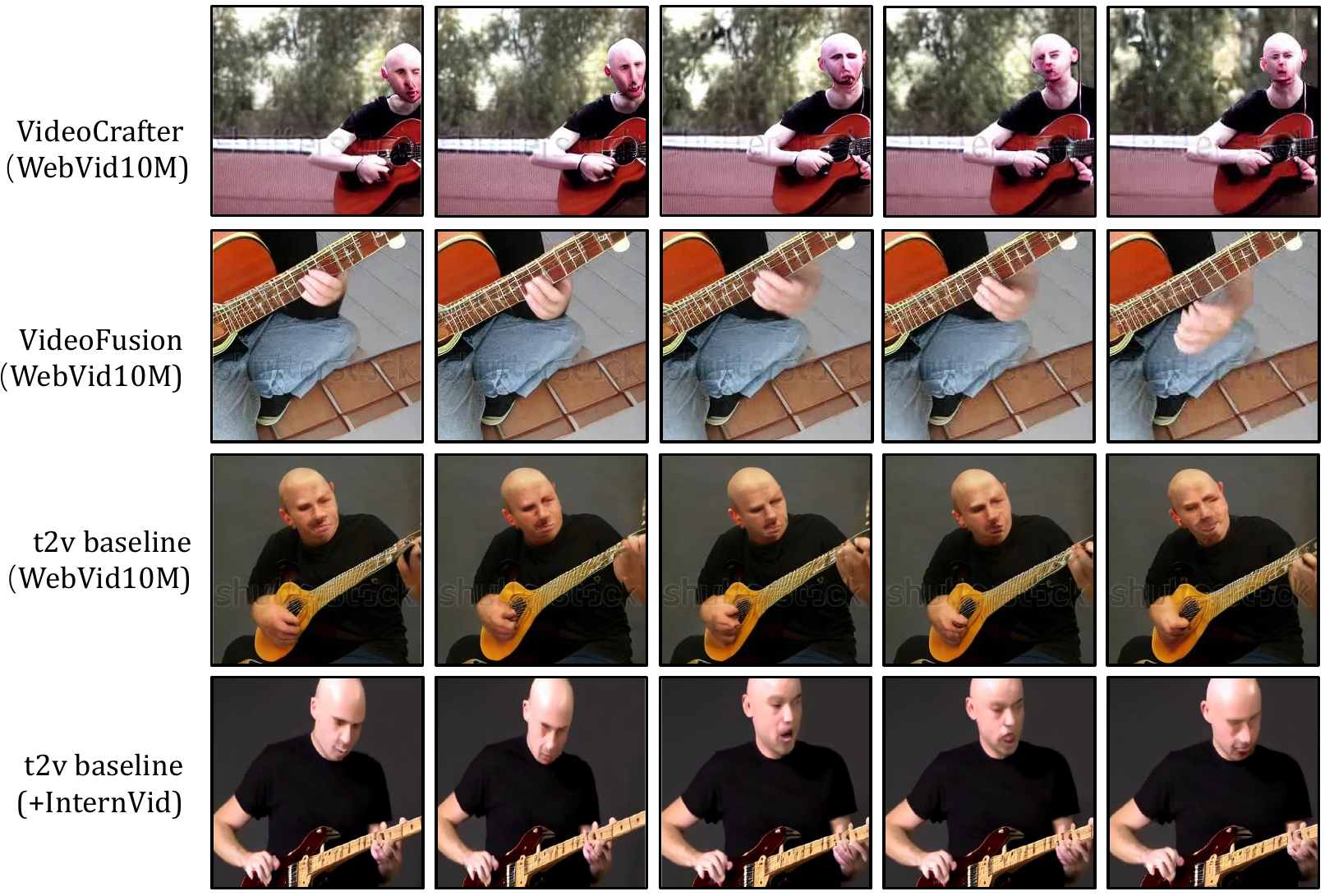}
    % \caption{Comparison of samples from t2v baseline to others. We provide zero-shot text-to-video generation results of different methods trained on both WebVid10M and the additional {\dataname}-Aes-18M. The used prompt is: \texttt{a bald man in a black t-shirt is playing a guitar.}
    % }
    \vspace{-0.3cm}
    \caption{Comparison of samples from t2v baseline to others. The used prompt is: \texttt{a bald man in a black t-shirt is playing a guitar.}
    }
    \label{fig:video_generation}
    \vspace{-0.5cm}
\end{figure*}

\subsection{Text-to-Video Generation}
% \vspace{-0.2cm}
Our {\dataname} dataset improves existing text-to-video generation models by providing video-text pairs with high correspondence. To establish a video generation baseline, we extend spatiotemporal modeling on the latent space of an open-source text-to-image diffusion model \cite{rombach2021highresolution}. We train the video generation approach with two settings: one using WebVid10M \cite{bain2021frozen}, and the other using {\dataname}-Aesthetics-18M in addition to WebVid10M \cite{bain2021frozen}. {\dataname}-Aesthetics-18M is a subset of {\dataname} consisting of clips with an aesthetic score of at least 4. Quantitative (Table \ref{tab:video_gen}) and qualitative (Figure \ref{fig:video_generation}) evaluations demonstrate the effectiveness of {\dataname} in video generation tasks. To evaluate our models quantitatively, we perform zero-shot text-to-video experiments and randomly sample 2,020 videos from the UCF-101 dataset and 2,990 videos from the MSRVTT dataset. Following the protocols in \cite{blattmann2023align}, we report CLIPSIM, IS, FID, and FVD metrics.

In Table \ref{tab:video_gen}, we observe that our t2v baseline trained on WebVid10M performs poorly in terms of IS, FID, and CLIPSIM when compared to other approaches. However, with the addition of {\dataname}-Aesthetics-18M, our t2v baseline demonstrates significant improvements in these metrics and outperforms other methods by a considerable margin. 
In Figure \ref{fig:video_generation}, we observe that the text-to-video (t2v) baseline using both WebVid10M and {\dataname}-Aesthetics-18M significantly outperforms other methods in terms of visual quality and temporal coherence. It is worth noting that the t2v baseline using {\dataname} does not contain watermarks, which is a data bias in WebVid10M. These results demonstrate the potential of {\dataname} for high-quality video generation.

\subsection{Video-Centric Dialogue System}

\begin{figure*}[t]
    \centering
    \includegraphics[width=1\textwidth]{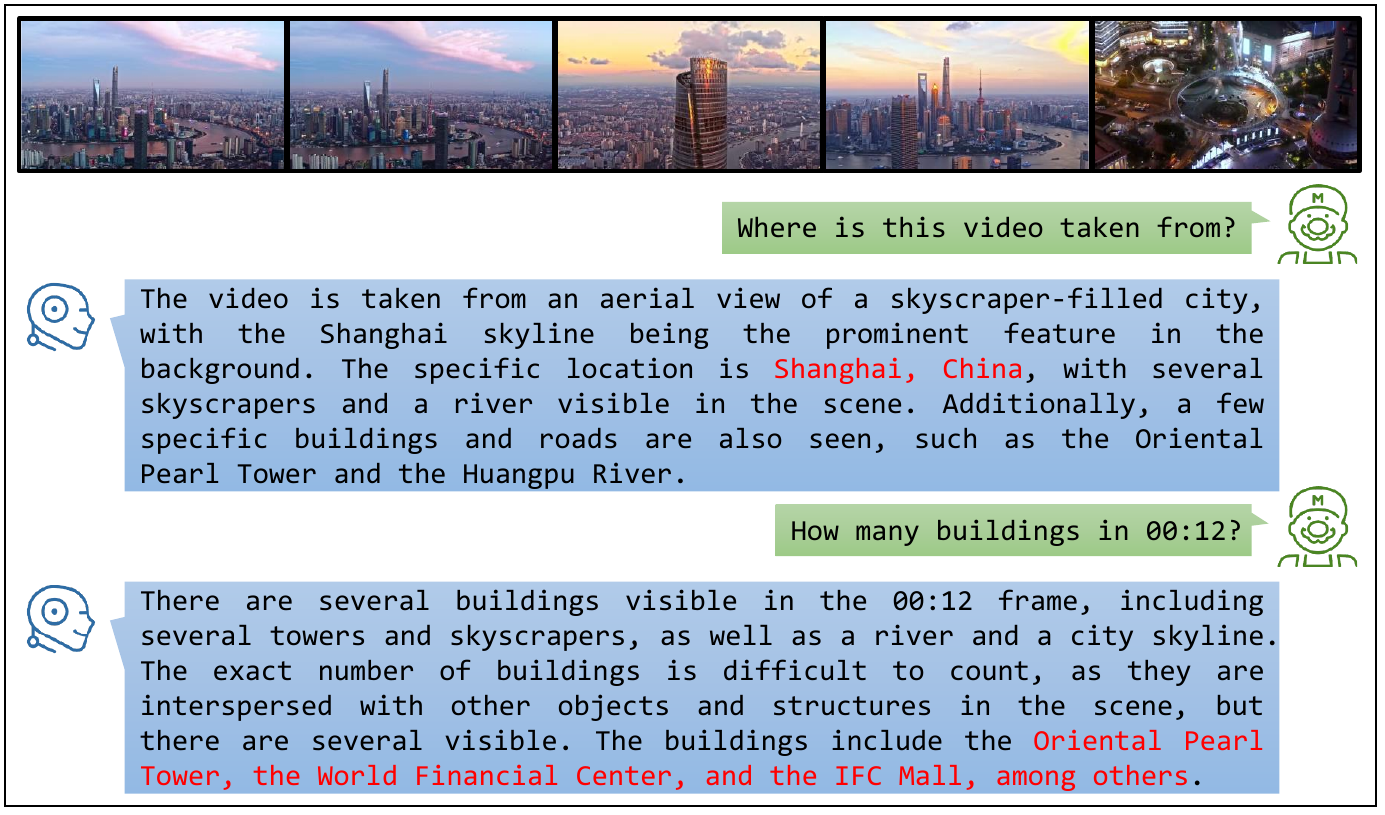}
    % \vspace{-0.3cm}
    \caption{\textbf{Video Spatial Understanding Task.} This figure demonstrates the spatial understanding and temporal localization capabilities of VideoChat-ViCLIP trained with our dataset.
    }
    \label{fig:vivo_spatial}
    \vspace{-0.3cm}
\end{figure*}

\begin{figure*}[t]
    \centering
    \includegraphics[width=1\textwidth]{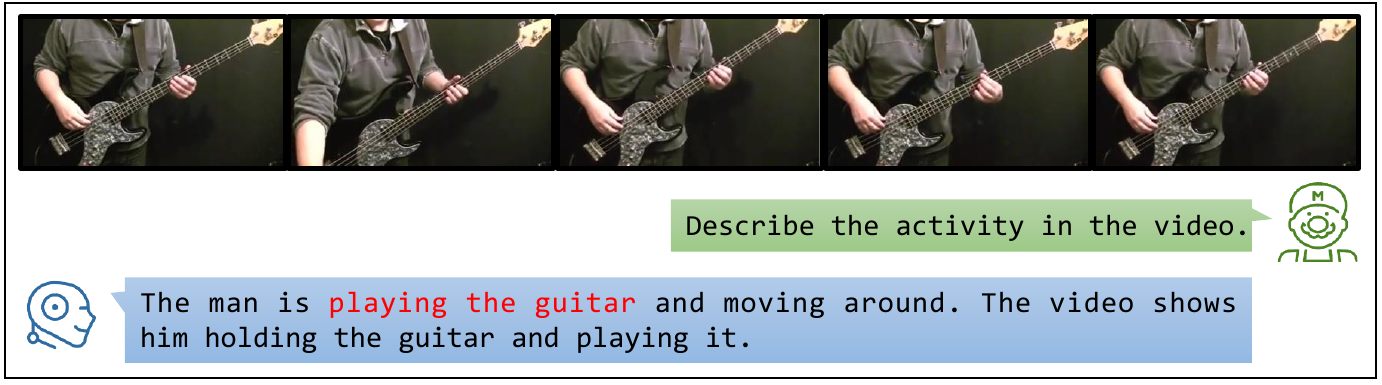}
    % \vspace{-0.3cm}
    \caption{\textbf{Video Action Recognition Task. } The video comes from Kinetics-400, with the label "playing guitar".}
    \label{fig:vivo_action}
    \vspace{-0.3cm}
\end{figure*}

\begin{figure*}[t]
    \centering
    \includegraphics[width=1\textwidth]{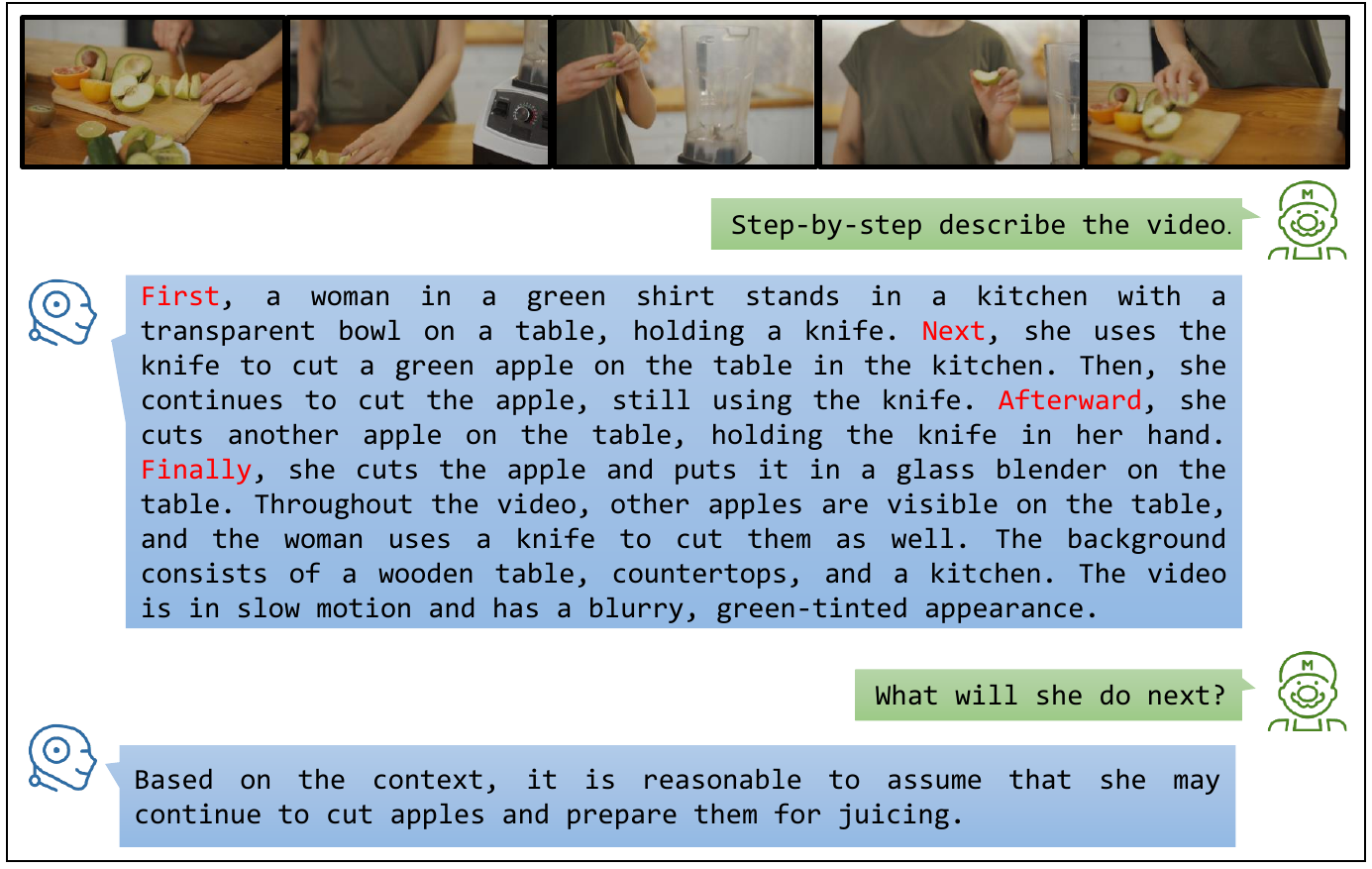}
    % \vspace{-0.3cm}
    \caption{\textbf{Temporal Understanding Task.} VideoChat-{\modelname} can easily handle temporal understanding tasks and make predictions based on these temporal sequences due to its training on rich temporal information on {\dataname}.
    }
    \label{fig:vivo_temporal}
    \vspace{-0.3cm}
\end{figure*}

\begin{figure*}[t]
    \centering
    \includegraphics[width=1\textwidth]{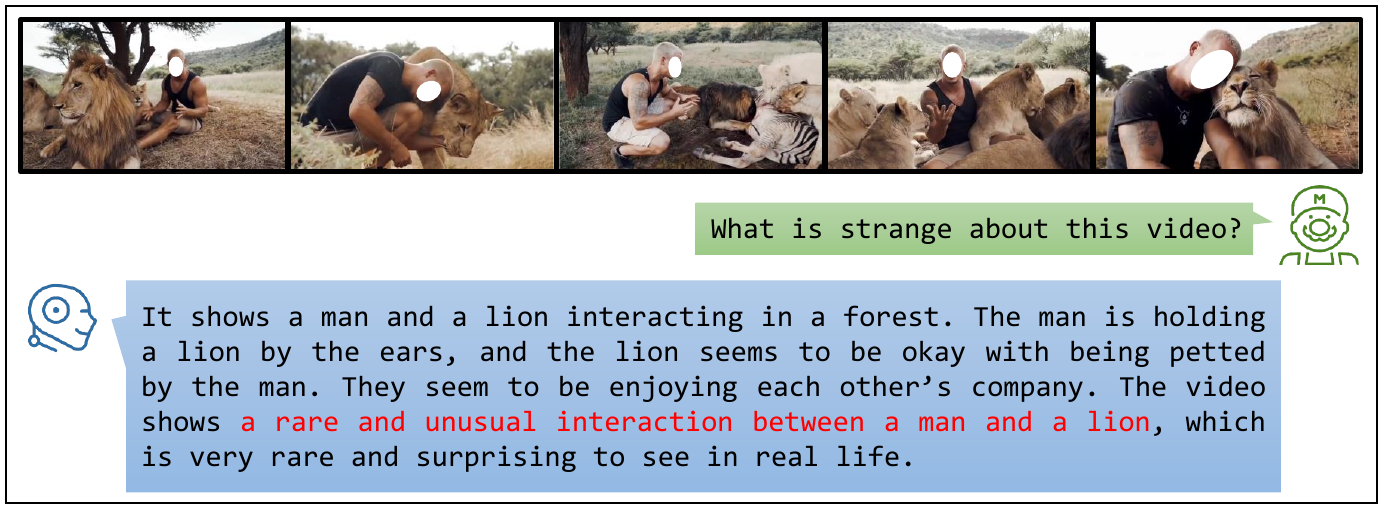}
    % \vspace{-0.3cm}
    \caption{\textbf{Video Reasoning Task.} Our dialogue system has the ability to understand common sense and make causal inferences based on the contents within the video. 
    }
    \label{fig:vivo_reasoning}
    \vspace{-0.3cm}
\end{figure*}

\begin{figure*}[th]
    \centering
    \includegraphics[width=1\textwidth]{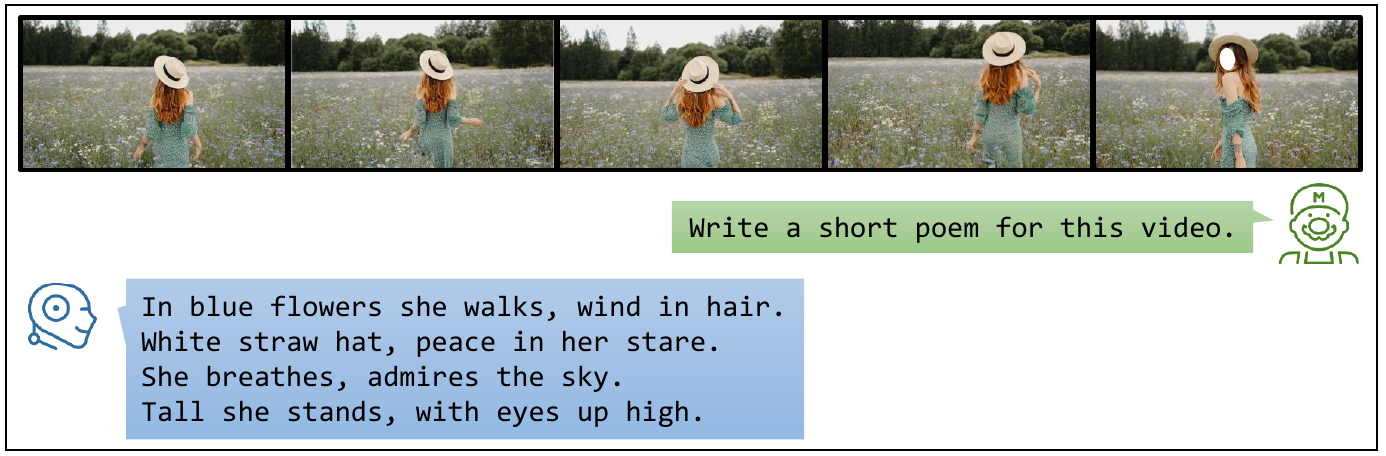}
    % \vspace{-0.3cm}
    \caption{\textbf{Video Creative tasks.} This figure shows the performance of VideoChat-ViCLIP on creative and generative tasks within our dataset, like generating poetry based on video content.}
    \label{fig:vivo_generate}
    \vspace{-0.3cm}
\end{figure*}

Inspired by recent vision-centric dialogue systems~\cite{li2023videochat,Maaz2023VideoChatGPT,li2023mimic}, we integrate our pretrained {\modelname} (with {\dataname}) into VideoChat~\cite{li2023videochat} to show how our data and model can empower multimodal dialogue methods with effective video modeling capability. In implementation, we inherit nearly all designs of VideoChat-Embed, just replacing its visual encoder with our {\modelname} (trained on {\dataname}). We evaluate VideoChat-{\modelname} in spatial understanding (Figure \ref{fig:vivo_spatial}), action recognition (Figure \ref{fig:vivo_action}), temporal understanding (Figure \ref{fig:vivo_temporal}), video reasoning (Figure \ref{fig:vivo_reasoning}), and video creative (Figure \ref{fig:vivo_generate}) tasks. Our qualitative evaluations demonstrate its decent video-to-text capabilities, suggesting promising potential for improving video captioning further. 

\begin{table}[th]
    \centering
    \resizebox{\textwidth}{!}{
    \begin{tabular}{l|lllll|l}
    \toprule
        % \multirow{2}{*}{Evaluation Aspect} & \multirow{2}{*}{Correctness of Information} & \multirow{2}{*}{Detail Orientation} & \multirow{2}{*}{Contextual Understanding} & \multirow{2}{*}{Temporal Understanding} & \multirow{2}{*}{Consistency} & \multirow{2}{*}{Avg} \\ %\hline
        \multirow{2}{*}{Evaluation Aspect} & Correctness & Detail & Contextual & Temporal & \multirow{2}{*}{Consistency} & \multirow{2}{*}{Avg} \\
        & of Information & Orientation & Understanding & Understanding & \\ \hline
        VideoChat (Eva-g) & 2.23 & 2.5 & 2.53 & 1.94 & 2.24 & 2.29 \\ %\hline
        LLaMA Adapter & 2.03 & 2.32 & 2.3 & 1.98 & 2.15 & 2.16 \\ %\hline
        Video LLaMA & 1.96 & 2.18 & 2.16 & 1.82 & 1.79 & 1.98 \\ %\hline
        Video-ChatGPT & 2.4 & 2.52 & 2.62 & 1.98 & 2.37 & 2.38 \\ %\hline
        VideoChat-ViCLIP & \textbf{2.86} & \textbf{2.52} & \textbf{3.08} & \textbf{2.36} & \textbf{2.4} &	\textbf{2.64} \\ 
        \bottomrule
    \end{tabular}
    }
    \caption{{Performance benchmarking of text generation models.}
    }
\end{table} \label{tab:textgen}

In terms of quantitative comparison, as shown in Table \ref{tab:textgen}, VideoChat-ViCLIP significantly outperforms the vanilla VideoChat (using Eva-g as the vision encoder) and other systems across all evaluation aspects of the quantitative video conversation evaluation framework in \cite{Maaz2023VideoChatGPT}. Specifically, the model shows remarkable improvements in the correctness of information (from 2.23 to 2.86), contextual understanding (from 2.53 to 3.08), and temporal understanding (from 1.94 to 2.36). The average score also increases from 2.29 to 2.64, showing an overall performance gain.

\section{Conclusion}
Our dataset, {\dataname}, is designed for multimodal research (both understanding and generation) focused on videos. It consists of over 230 million video clips sourced from 7 million high-resolution (720P) YouTube videos. We use existing models with a multiscale approach to generate clip-level descriptions. Our studies confirm the efficacy of captions, and the large volume of video-text data enables crossmodal learning and text-to-video generation at scale. By training with our data, we develop a video-text representation baseline {\modelname} using ViT-L and analyze briefly how the data scale affects learned crossmodal embeddings. In addition to perception tasks, we show that {\dataname} improves text-to-video generation performance when using a subset of clips based on their aesthetic scores. With its data, annotations, metadata, and computed scores, we believe {\dataname} can fuel a variety of studies and applications.

\appendix

\begin{figure}[h]
    \centering
    \includegraphics[width=1\textwidth]{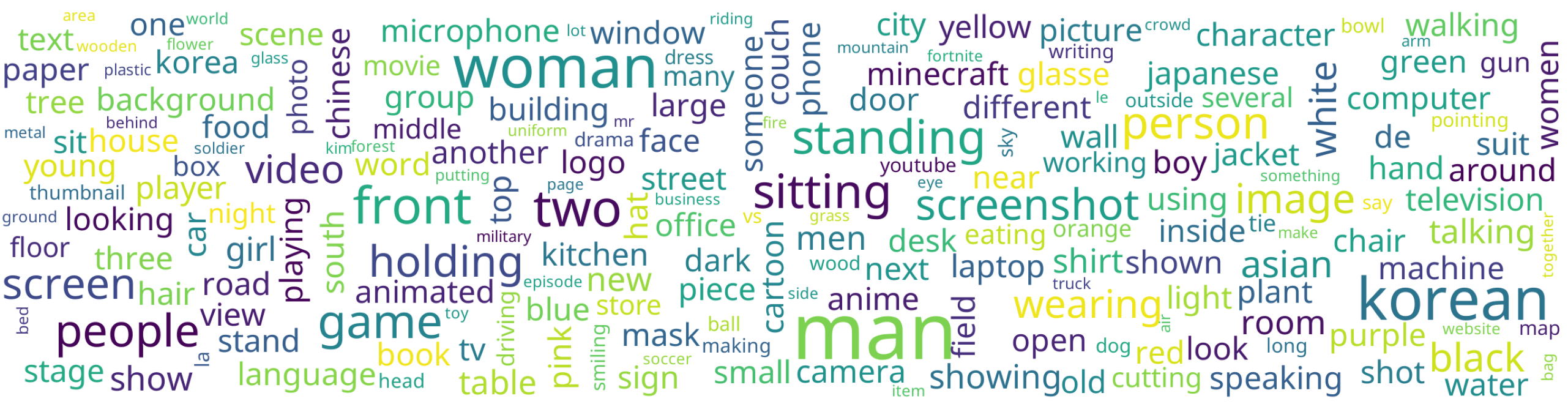}
    \caption{The word cloud (Top-200) of the generated captions in the {\dataname} dataset reveals that the captions predominantly highlight the rich actions of the objects.}
    \label{fig:word_cloud_caption}
\end{figure}

\begin{figure}[t]
  \centering
  \begin{minipage}[b]{0.45\linewidth}
    \centering
    \includegraphics[width=\linewidth]{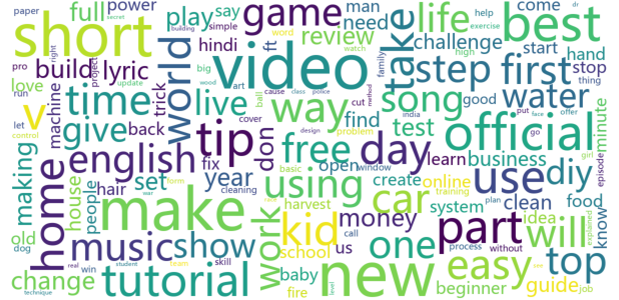}
    English.
  \end{minipage}
  \hfill
  \begin{minipage}[b]{0.45\linewidth}
    \centering
    \includegraphics[width=\linewidth]{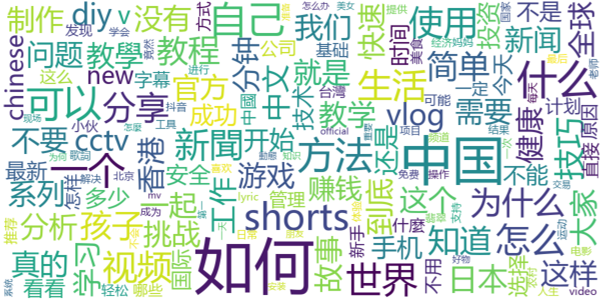}
    Chinese.
  \end{minipage}
  \vskip\baselineskip
  \begin{minipage}[b]{0.45\linewidth}
    \centering
    \includegraphics[width=\linewidth]{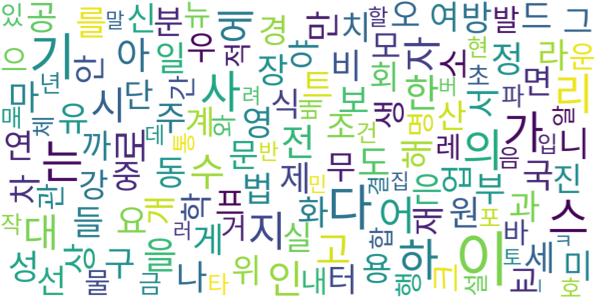}
    Korean.
  \end{minipage}
  \hfill
  \begin{minipage}[b]{0.45\linewidth}
    \centering
    \includegraphics[width=\linewidth]{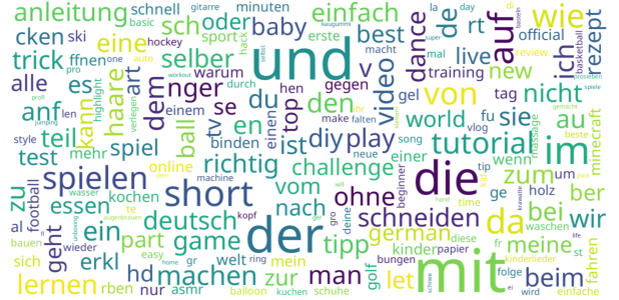}
    German.
  \end{minipage}
  \caption{The word clouds of the ASR transcripts of four different languages (English, Chinese, Korean, and German). We collect videos from various countries or regions with 11 different languages. Here we list four of them to show how these transcripts are distributed in words.}
  \label{fig:worddist}
\end{figure}

\section{Data Availability Statement} \label{sec:data_ava}
We are committed to maintaining transparency and compliance in our data collection and sharing methods. In accordance with these principles, please note the following:

\textbf{Publicly Available Data:} The data utilized in our studies is publicly available. We do not use any exclusive or private data sources.

\textbf{Data Sharing Policy:} Our data sharing policy builds upon the precedent set by prior works like Kinetics, HD-VILA, and others. Instead of providing the original raw data, we only supply the YouTube video IDs necessary for downloading the respective content.

\textbf{Usage Rights:} The data released by us is intended exclusively for research purposes. Any potential commercial usage is not sanctioned under this agreement.

\textbf{Compliance with YouTube Policies:} Our data collection and release practices are strictly in accord with YouTube's data privacy policies. We ensure that no user data or privacy rights are violated during the process.

\textbf{Data Licence:} We employ the protocol of CC BY 4.0.

\section{Limitations \& Societal Impact}
All video data used in our research are downloaded from YouTube using Safe for Work (SFW) queries and channels. To ensure appropriate content, we employ a simple NSFW filter: a binary classifier designed to recognize and exclude non-ethical videos. For privacy considerations and in respect of data sharing practices, we share only the YouTube ID of the videos, similar to previous academic works. This approach aligns with YouTube's data protocols and ensures no violation of privacy or data usage rules. Despite these precautions, our work has some limitations, primarily related to data diversity and representativeness. Although YouTube is an extensive source encompassing a wide range of video categories, certain specific types of footage may be excluded or scarcely collected, including: public area surveillance, sports competitions, movies, documentaries, etc. The exclusion of such categories is often due to copyright restrictions or other limits imposed by the platform. Therefore, while our dataset provides a broad view of everyday video content, its coverage does not extend to every possible category or type of video. These limitations should be taken into account when considering the generalizability of our results across all types of video data.

\section{More Statistics in {\dataname}} \label{sec:stat_caps}
% \subsection{}
% \subsection{Statistics of {\dataname}}
\paragraph{Actionness.}
{\dataname} contains way more verbs than the WebVid10M. We used NLTK toolkit to analyze the number of verbs in captions, focusing on tagging all unique verbs. We found a total of 109,485 verbs in the WebVid10M, while {\dataname} contained 212,155 ones. While the counts may not be that accurate due to our simple counting, we believe they provide a rough indication of the actionness of the two datasets.

\paragraph{Video Caption and Transcript Distribution.} To analyze the word distribution of our generated captions and multilingual (ASR) transcripts, we compute their distributions. The resulting word distribution of the captions is presented in Figure \ref{fig:word_cloud_caption}, which includes objects (tv, car, door, plant, etc.), attributes (green, young, large, long, etc.), locations (middle, behind, south, next, etc.), scenes (room, stage, kitchen, office, etc.), actions/events (walking, eating, cutting, holding, etc.), and more.

We also include four word distributions of different languages in Figure \ref{fig:worddist}, reflecting trends in different countries and offering potential data customization along with the provided metadata.

% \end{document}

\section{{\dataname}-ICL: Interleaved Video-Text for In-Context Video Learning}

\begin{table*}[h!]\centering
    \begin{minipage}{0.99\columnwidth}\vspace{0mm}    
    \centering
    \begin{tcolorbox} 
        \centering
        \hspace{-6mm}
        \begin{tabular}{p{0.99\columnwidth}}
        \hspace{1mm}
        \begin{minipage}{0.99\columnwidth}

        \texttt{[..., "the inside of a home has a rug and a light on.", \textcolor{gray}{"♪ We could leave the Christmas lights up til January ♪"}, ..., "woman with blond hair playing guitar", \textcolor{gray}{"♪ Have I known you 20 seconds or 20 years? ♪",}}
        {\makebox[0pt][l]{\hspace{0pt}\raisebox{-0.3ex}{{\includegraphics[height=25pt]{supp/case1-1.pdf}}}}}
        \texttt{~~~~~~~~~~~~~~~~~~~~~~~~~~~~, "close-up of a bathroom sink with soap bubbles and other items", "a bathroom is seen with a sink and two lights", "a woman swiming inside of a fishbowl with a ladder and a man", \textcolor{gray}{"♪ Can I go wher you go? ♪",}}
        {\makebox[0pt][l]{\hspace{0pt}\raisebox{-0.3ex}{{\includegraphics[height=25pt]{supp/case1-3.pdf}}}}}
        
        \texttt{, "devils roll the dice, angels roll their eyes",\textcolor{gray}{"♪ And, take me out, and take me home ♪" ,}..., "the man is standing in a room with pink carpet",\textcolor{gray}{"♪ You're my, my ♪"}, "a woman in yellow is dancing with a man in a red room", \textcolor{gray}{"♪ My, My lover ♪",} }\\       
        {\makebox[0pt][l]{\hspace{0pt}\raisebox{-0.3ex}{{\includegraphics[height=25pt]{supp/case1-4.pdf}}}}}
        \texttt{~~~~~~~~~~~~~~~~~~~~~~~~~, "a woman is sitting on a chair, playing a guitar and a woman holding a balloon", \textcolor{gray}{"♪ ♪ ♪",} "two men smiling while holding wine glasses and drinking beer", \textcolor{gray}{"♪ We could let our friends crash in the living room ♪"} ...]}
        \end{minipage}
        \end{tabular}
    \end{tcolorbox}
    \vspace{-2mm}
    \caption{\textbf{Interleaved video-text data format (b) in {\dataname}.} The caption and ASR transcript of each clip is shown in black and gray, respectively. We can achieve interleaved video-text data format (a) by abandoning ASR transcripts. To obtain data format (c), we concatenate multiple videos with interleaved video-text data (a).
}
    \label{tab:video-icl1}
\end{minipage}
    \vspace{-2mm}
\end{table*}

\paragraph{Visual Examples.} As given in the paper, we provide examples video+text interleaved entries for in-cntext learning as Flamingo. Table \ref{tab:video-icl1} gives an example about format (a): arrange clips and their descriptions sequentially based on their temporal order within the same video. Note the videos are randomly dropped with a probability (0.3) for constructing richer text context compared with the original video-text pair combinations in sequential.

\section{Implementation Details}
\subsection{{\modelname}}  \label{sec:vclip_set}
\paragraph{Action Recognition.} In the zero-shot action recognition, we sample 8 frames in each video.
Following the settings in CLIP and EVA-CLIP, we report the mean of top-1 and top-5 accuracy for Kinetics-400 / -600 / -700. In Section \ref{sec:zs-action}, we show {\modelname} learnt on WebVid or {\dataname} is an 
effective zero-shot action recognition model.

In the full fine-tuned setting, we conduct two experiments with two receipts. In Table \ref{tab:ft_recognition}, for the experiments where the training data excluded K710, we followed the common practice of finetuning the pretrained {\modelname} with the training data from the evaluation dataset. On the other hand, for the experiments where the training data included K710, we adopted a training trick inspired by \citep{uniformerv2}. We first finetuned the pretrained {\modelname} with K710 \citep{uniformerv2}, and then proceeded with the common supervised finetuning setting. By incorporating the supervised finetuning with K710, {\modelname} demonstrated better performance in the fine-tuned tasks compared to experiments that did not include K710.

\paragraph{Video Retrieval.}
In the full-finetuning setting, we tune the pretrained {\modelname} with not only video-text contrastive loss but also video-text matching loss on the training data of the evaluated benchmarks. During both training and testing, we sample 12 frames. Detailed hyper-parameters are given in Table \ref{tab:ret_hyperparameters}. In the zero-shot setting, we sample only 8 frames for evaluations.

\begin{table*}[t!]
    \centering
    \setlength\tabcolsep{4pt}
    % \resizebox{1.0\linewidth}{!}{
        \begin{tabular}{l|ccccc}
        config & MSRVTT & DiDeMo & ANet & LSMDC & MSVD \\
        \Xhline{1.0pt}
        optimizer & \multicolumn{5}{c}{AdamW} \\ 
        optimizer momentum & \multicolumn{5}{c}{$\beta_1, \beta_2{=}0.9, 0.999$}  \\
        weight decay & \multicolumn{5}{c}{0.02} \\
        learning rate schedule & \multicolumn{5}{c}{cosine decay} \\
        % learning rate & 2e-5 (B/L) & 2e-5 (B), 4e-5 (L) & 4e-5 (B/L) & 2e-5 (B/L) & 2e-5 (B/L) \\
        learning rate & 2e-5 & 4e-5 & 2e-5 & 2e-5 & 4e-5 \\
        batch size & \multicolumn{5}{c}{256} \\
        warmup epochs & \multicolumn{5}{c}{1} \\
        % total epochs &  10 (B), 7(L) & 12 (B), 5 (L) & 20 (B/L) & 10 (B), 8 (L) & 10 (B/L)  \\
        total epochs &  7 & 8 & 5 & 10 & 20  \\
        input frame & \multicolumn{5}{c}{12} \\
        % max text length & 32 & 64 & 150 & 96 & 64 \\
        max text length & 32 & 96 & 64 & 64 & 150 \\
        % drop path & 0.2 (B), 0.3 (L) & 0.1 (B), 0.3 (L) & 0.1 (B), 0.2 (L) & 0.1 (B), 0.2 (L) & 0.2 (B), 0.3 (L) \\
        drop path & 0.3 & 0.2 & 0.3 & 0.3 & 0.2 \\
        flip augmentation & \multicolumn{5}{c}{\textit{yes}} \\
        augmentation & \multicolumn{5}{c}{MultiScaleCrop [0.5, 1]} \\
        \end{tabular}
    % }
    \caption{
        \textbf{Video-text retrieval fine-tuning settings.}
    }
    \vspace{-0.3cm}
    \label{tab:ret_hyperparameters} 
\end{table*}

\subsection{Video Generation Baseline} \label{sec:vid_gen_baseline}

\begin{figure*}[t]
    \centering
    \includegraphics[width=1\textwidth]{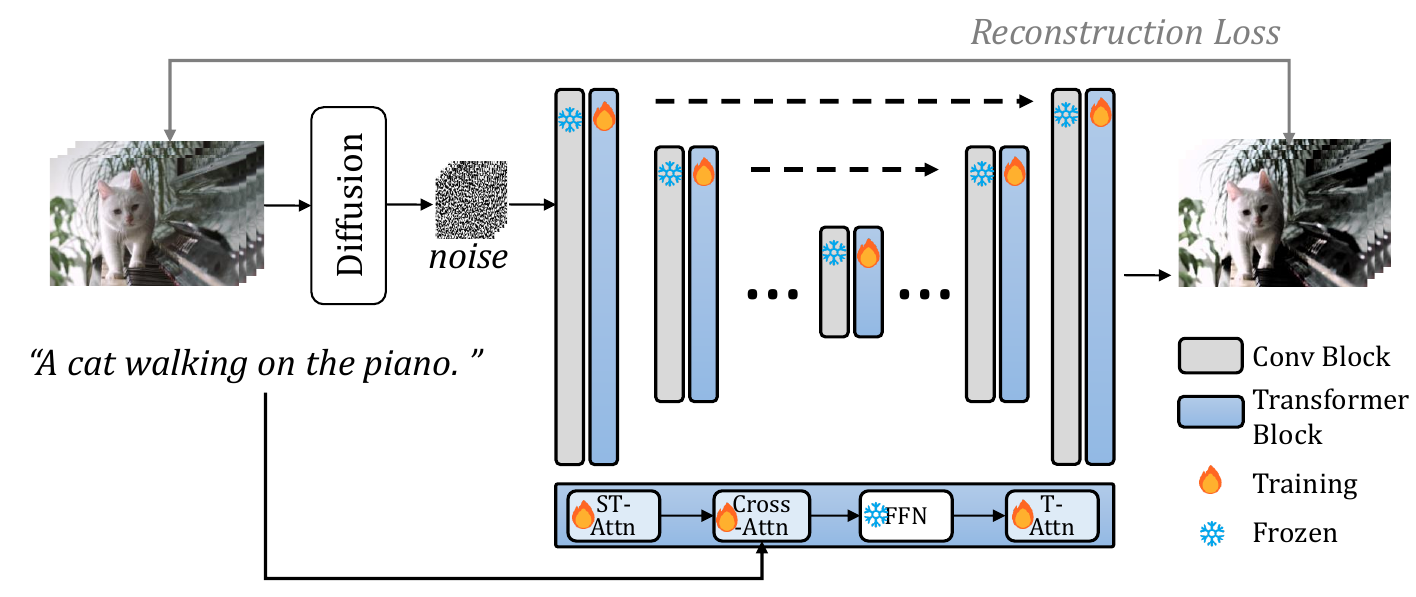}
    % \vspace{-0.3cm}
    \caption{Framework of our text-to-video generation baseline.}
    \label{fig:generation_pipeline}
    \vspace{-0.3cm}
\end{figure*}

We used the spatiotemporal modeling approach from \citep{wu2022tune} and built our text-to-video generation baseline on the work of \citep{rombach2021highresolution}. Our approach consists of a U-Net with a transformer that models its latents, using interleaved spatiotemporal attention (ST-Attn), cross-attention for visual-text, a feed-forward network (FFN), and temporal attention (T-Attn), as illustrated in Figure \ref{fig:generation_pipeline}. To adapt the 2D convolutional layers in \citep{rombach2021highresolution} to 3D, we extended $3 \times 3$ kernels into $1 \times 3 \times 3$ ones. We also extended the original spatial attentions to spatiotemporal ones. We initialized our baseline using all text-to-image diffusion model parameters, while the newly added temporal attention layers used default parameters.

For the ST-Attn implementation, we used frame embeddings from the U-Net encoder instead of video embeddings as in \citep{wu2022tune}. We concatenated the embeddings of the previous and current frame for values and keys in attention, while using the current frame embedding alone as queries. The rest of the implementation remained the same as the original.

% \paragraph{Full Model Hyper-parameters.}
\paragraph{Text-to-Video Evaluation.}
To evaluate our text-to-video model, we conducted zero-shot experiments on the UCF-101 and MSRVTT datasets, following the method from \citep{blattmann2023align}. For UCF-101, we used the class names as text prompts and generated 20 samples per class (total of 2,020 videos). For MSRVTT, we randomly selected one caption per video from the official test set (total of 2,990 videos). To ensure a fair comparison, we used the official implementation of VideoCrafter and VideoFusion \citep{luo2023videofusion} to generate the same number of videos with the same text prompts. During video sampling and evaluation, we generated 16 frames per video.

We assess the overall quality of the synthesized results on UCF-101 using framewise-FID, FVD, and Inception Score (IS), and evaluate the text-video semantic similarity on MSRVTT using clip similarity (CLIPSIM). For framewise-FID and IS, we use the pretrained Inceptionv3 network weights as our image encoder. For FVD, we use the pretrained InceptionI3d model and followed the TATS method \citep{ge2022long}. To compute CLIPSIM, we calculate the clip text-image similarity for each frame with respect to the given text prompts and computed the average score. We use the ViT-B-32 clip model as the backbone, consistent with previous work \citep{blattmann2023align}.

\section{More Results}

\begin{table}[t]
% \scriptsize
\centering
% \resizebox{\textwidth}{!}{
% \scriptsize
% \small
\setlength{\tabcolsep}{2.0 mm}{
\begin{tabular}{@{}ccccccccccc@{}}
\toprule
\multirow{4}{*}{\begin{tabular}[c]{@{}c@{}}Captioning \\ Method\end{tabular}} & \multicolumn{4}{c}{Retrieval} & \multicolumn{6}{c}{Action Recognition} \\
 & \multicolumn{2}{c}{Zero-Shot} & \multicolumn{2}{c}{Fine-Tuned} & \multicolumn{6}{c}{Zero-Shot} \\
 & \multicolumn{2}{c}{MSR-VTT} & \multicolumn{2}{c}{MSR-VTT} & \multicolumn{2}{c}{K400} & \multicolumn{2}{c}{K600} & \multicolumn{2}{c}{K700} \\
 & T2V & V2T & T2V & V2T & top-1 & AVG & top-1 & AVG & top-1 & AVG \\
 \midrule
VideoChat & 33.9 & 32.3 & 46.6 & 47.1 & 54.68 & 67.74 & 51.70 & 64.91 & 43.67 & 56.51 \\
Ours & \textbf{38.6} & \textbf{38.5} & \textbf{49.0} & \textbf{49.2} & \textbf{58.52} & \textbf{71.11} & \textbf{55.37} & \textbf{68.27} & \textbf{47.09} & \textbf{59.98} \\
\bottomrule
\end{tabular}
}
% }
% \vspace{-3mm}
\caption{Video retrieval and action recognition results of {\modelname}-B trained on {\dataname}-FLT-10M with the captions generated by VideoChat and our captioning approach.}
% \vspace{-3mm}
\label{tab:diff_cap}
\end{table}

\subsection{Effectiveness of Our Multiscale Captioning Approach}
To further validate the effectiveness of our proposed captioning method, we establish a video caption baseline using the video multimodal model VideoChat \citep{li2023videochat} for comparison. We input the video clip into the model with the prompt \texttt{"Please describe the content in the given video."} and apply it to {\dataname}-10M-FLT, resulting in 10 million new captions generated by VideoChat. Subsequently, we train two versions of {\modelname}-Base using {\dataname}-10M-FLT, each version trained with one of the two types of captions.

Table \ref{tab:diff_cap} demonstrates that {\modelname}-B trained using our captions outperforms the version trained using captions from VideoChat in both video retrieval (MSR-VTT) and action recognition (K400/600/700). These results are particularly noteworthy considering that the only difference in training lies in the captions generated by the two different approaches. Therefore, these findings further confirm the superior performance of our proposed captioning method compared to the baseline VideoChat.

\begin{figure*}[t]
    \centering
    \includegraphics[width=0.95\textwidth]{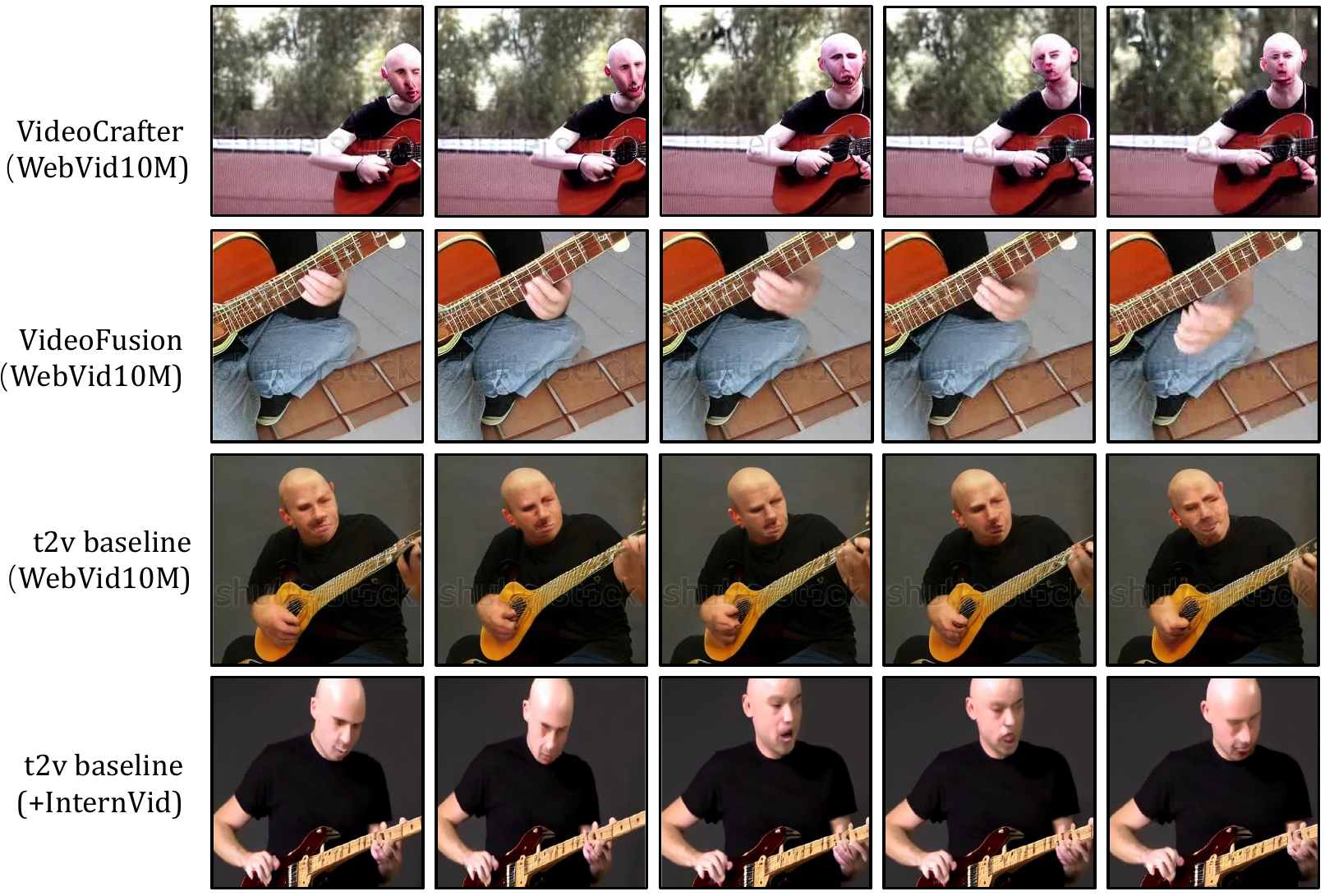}
    \caption{Comparison of samples from t2v baseline to others. We provide zero-shot text-to-video generation results of different methods trained on both WebVid10M and the additional {\dataname}-Aes-18M. The used prompt is: \texttt{a bald man in a black t-shirt is playing a guitar.}
    % }
    \vspace{-0.3cm}
    % \caption{Comparison of samples from t2v baseline to others. The used prompt is: \texttt{a bald man in a black t-shirt is playing a guitar.}
    }
    \label{fig:video_generation}
\end{figure*}

% \subsection{Temporal Action Localization}
% We evaluate {\modelname} and OpenAI's CLIP on a temporal action localization dataset Thumos, using the task head Actionformer with the full-finetuning setting. 

\subsection{Text-to-Video Generation}
In Figure \ref{fig:video_generation}, we observe that the t2v baseline using both WebVid10M and {\dataname}-Aes-18M significantly outperforms others in visual quality and temporal coherence. 
Note that the t2v baseline using {\dataname} does not contain watermarks, which is a data bias in WebVid10M. 

{\small
\bibliographystyle{unsrt}
\bibliography{egbib}
}

\end{document}